\documentclass[
    aip,
    rsi,
    reprint
]{elsarticle}
\usepackage{lineno}
\modulolinenumbers[5]

\journal{Physica D}
\usepackage[obeyDraft,obeyFinal]{todonotes}
\usepackage{amsmath}

\usepackage{amssymb}
\usepackage{graphicx}

\usepackage{algorithm}
\usepackage{algpseudocode}
\usepackage{booktabs}
\usepackage{overpic} 

\def\R{\mathbb{R}}
\newcommand{\secref}[1]{Sec.~\ref{#1}} 
\newcommand{\tabref}[1]{Tab.~\ref{#1}} 
\newcommand{\figref}[1]{Fig.~\ref{#1}} 

\newcommand{\mc}[1]{\multicolumn{1}{c}{#1}}  

\begin{document}

\begin{frontmatter}
	
	\title{Explainable Machine Learning Control - robust control and stability  analysis}
	%
	\author{Markus Quade}
	\address{Ambrosys GmbH, David-Gilly Stra{\ss}e 1, 14469 Potsdam, Germany}
	\author{Thomas Isele}
	\address{4Cast GmbH \& Co. KG, Parkstra{\ss}e 1, 14469 Potsdam, Germany}
	\author{Markus Abel\corref{mycorrespondingauthor}}
	\address{Ambrosys GmbH, David-Gilly Stra{\ss}e 1, 14469 Potsdam, Germany}
	\address{4Cast GmbH \& Co. KG, Parkstra{\ss}e 1, 14469 Potsdam, Germany}
	\address{Universit\"at Potsdam, Institut f\"ur Physik und Astronomie, Karl-Liebknecht-Stra{\ss}e 24/25, 14476 Potsdam, Germany}
	\cortext[mycorrespondingauthor]{Corresponding author}
	\ead{markus.abel@ambrosys.de}

	\begin{abstract}
		Recently, the term explainable AI became known as an approach to produce models from artificial intelligence which allow interpretation. Since a long time, there are models of symbolic regression in use that are perfectly explainable and mathematically tractable:
		in this contribution we demonstrate how to use symbolic regression methods to infer the optimal control of a dynamical system given one or several optimization criteria, or cost functions. In previous publications, network control was achieved by automatized machine learning control using genetic programming. Here, we focus on the subsequent analysis of the analytical expressions which result from the machine learning. In particular, we use AUTO to analyze the stability  properties of the controlled oscillator system which served as our model. As a result, we show that there is a considerable advantage of explainable models over less accessible neural networks.
	\end{abstract}
	
	\begin{keyword}
		Explainable AI \sep Machine Learning Control \sep Dynamical systems \sep Synchronization Control \sep Genetic programming
	\end{keyword}
	
\end{frontmatter}
%
%
%
%
%
%
%
%
\section{Introduction}
Machine learning and artificial intelligence have recently rediscovered so-called explainable methods \cite{10.1007/978-3-319-99740-7_21}. Whereas this sounds appealing, researchers agree that explainability and interpretability is neither a new concept nor new in artificial intelligence or machine learning. The wish for it arose in recent years with the understanding that the very successful methods of deep learning with neural networks are not  directly interpretable. On the other hand, symbolic regression methods are not so new but very explainable, in particular genetic programming methods are extremely general, but their convergence and solutions are not as performant as are neural network methods. Generalized regression methods, on the other hand are very performant, but not so flexible. Here, we demonstrate the power such models reveal by extending a previous analysis of network machine learning control by a subsequent stability analysis. 

As an example for the control of dynamical systems in physics \cite{PhysRevLett.64.1196,chen2003chaos} or medicine \cite{haken2006brain,schwalb2008history} we chose the control of synchronization. Synchronization is a widespread phenomenon observed in many natural and engineered complex systems whereby locally interacting components of a complex system tend to coordinate and exhibit collective behavior \cite{Pikovsky2003Synchronization,strogatz2003sync}.
In \cite{Gout2018} synchronization in networks is investigated by multiple weakly coupled independent oscillating systems; the control then influences the overall dynamics of the system. The role control is to  drive the system into or out of synchronization by applying an external control signal \cite{Pikovsky2003Synchronization} that in turn depends on the state itself. The realization of this technique resembles reinforcement learning and differences and similarities are discussed elsewhere.
There are significant implications for numerous domains in engineering and science, including communications \cite{yang1997impulsive}, teleoperations \cite{li2011adaptive,shokri2014optimal} and brain modeling \cite{hammond2007pathological}. The special topic, especially phase oscillators, is reviewed in \cite{dorfler2013synchronization}.
Depending on the system, control may be those based on control theory \cite{kirk2012optimal}, mathematical and numerical optimization \cite{nocedal2006numerical} and computational intelligence \cite{opt4ml} techniques. The ``optimal control'' methods \cite{Becerra2008OptimalControl} aim at driving and maintaining a dynamical system in a desired state. This is typically achieved by finding a control law, in the form of a set of differential equations, which optimizes (by maximizing or minimizing) a cost function related to the control task. If the control is useful is decided heavily by the stability of the controlled system. 

This stability can be determined by standard mathematical methods for linear theory can be used \cite{Kirk1970OptimalControl,scholl2008handbook}. However, for nonlinear, extended and consequently complex systems, linear theory to determine a control may fail. In such cases, the more general methods used here can be of use, where analytical expressions are determined in a deterministic or evolutionary way. I.e. control laws are inferred from an arbitrary domainusing evolutionary machine learning methods as a suitable source of algorithms. Specifically, we refer to genetic programming (GP) \cite{Koza1992GeneticProgramming} to control synchronization in coupled networks, including a hierarchical network of coupled oscillators. Unlike neural networks and other black-box artificial intelligence methods, GP allows dynamically learning complex control laws in an interpretable symbolic form --- a method that is referred as symbolic regression \cite{schmidt2009distilling,vladislavleva2009order,Quade2016Prediction}. In particular, we focus on the subsequent rigorous analysis of the optimal laws found.
In contrast to previous works a full expression is optimized, and not only parameters \cite{shokri2014optimal,shokri2014comparison}. 

Based on the previous results, we demonstrate the effectiveness of the control and rigorous analysis exemplary by the analysis of control solutions for two oscillators found by symbolic regression to synchronize or de-synchronize the oscillators. Further application to the control of entire networks is straightforward and is to be included in the future in the fully automatized software framework Glyph \cite{,markus_quade_2017_801819}. The motivation is not only motivated by brain disorder problems in the medical domain like body tremors occur when firing neurons synchronize in regions of brain \cite{haken2006brain}, but may find broad application in any control setup, e.g. in machinery \cite{doi:10.2514/6.2018-3684}. In the case of brain states, if the firing of neurons is periodic, which may appear due to the inherent dynamics of the excitable neurons, a mutual influence may give rise to synchronization \cite{Pikovsky2003Synchronization,Strogatz2006SyncBasin}. If the coupling term is very large, this synchronization may extend over a whole region in our brain and thus over many neurons. Eventually this collective firing leads to shaky movements of hands, arms or the head, and is treated as a brain disorder. One remedy to this problem is to implant a control device which resets the neurons and counteracts the collective synchronization. An evident question then is how to design such a controller which also minimizes design cost, energy consumption, or other medical constraints but is as stable and reliable as possible. We analyze the found control technically by the well-known package AUTO \cite{AUTO07p}. 

To the best of our knowledge, this is the is first demonstration of machine learning control followed by automatized stability analysis. The extension and generalization is straightforward subject of an extension of existing software. GP is used to learn a control that brings a network of self-sustained oscillators in a desired, synchronized or de-synchronized state and back. In Sec. \ref{sec:results}, the results of our study of GP application to networked dynamic systems is presented with focus on the stability analysis. 

This publication is structured as follows: in Sec.~\ref{sec:methods} we recall the methods used and previous insights, in Sec.~\ref{sec:results} we reiterate results for machine learning control and discuss in great detail the stability analysis and thereby robustness of control for an exemplary control term, the publication ends with a discussion and conclusion in Sec.~\ref{sec:conc}. In the appendix we provide details on the parameters used for system integration and GP setup.

\section{Methods}
\label{sec:methods}

In this section, we briefly recall the concept of MLC (machine learning control) and the method used here to solve the problem.To control a dynamical system, one determines a manipulation of the trajectory of the system in phase space to drive it to and keep it in a desired state. This control problem is typically formulated as an optimization problem with an objective function that is to be minimized. In the control setup, this objective is formulated as the deviation of the state of the system from its desired one.
Consequently, an optimal control problem is formulated as a mathematical model of the system, a cost function or performance index, a specification of boundary conditions on states, and additional constraints. According to the type of problem, it is classified roughly according to Fig.~\ref{fig:optimal-control}

\subsection{Machine Learning Control \label{sec:feedback-control}}
\label{sec:oc}

\begin{figure}[htpb]
	\centering
	\includegraphics[width=1.\textwidth]{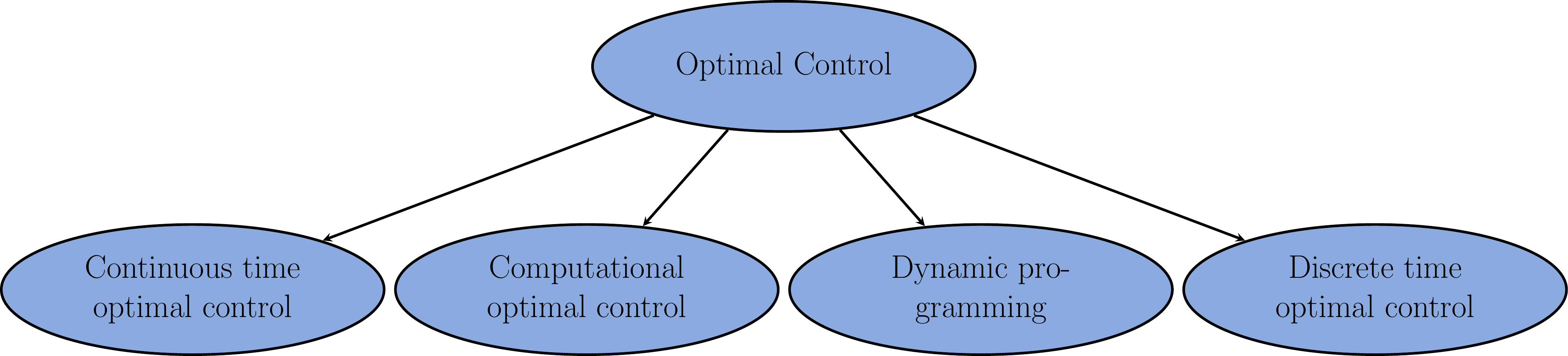}
	\caption{An overview of different types of optimal control. \label{fig:optimal-control}}
\end{figure}

Here, we focus on continuous-time optimal control. If there are no constraints on the states or the control variables, and if the initial and final conditions are fixed, it reads: Find the control vector $\vec{u}:\R^{n_x} \times [t_s, t_f] \mapsto \R^{n_u}$ that minimizes the cost function
\begin{align}
\Gamma = \varphi(\vec{x}(t_f)) + \int_{t_s}^{t_f} L(\vec{x}(t), \vec{u}(\vec{x}, t), t) \mathrm{d}t , \label{eq:oc-gamma}
\end{align}
subject to
\begin{align}
\dot{\vec{x}} = \vec{\tilde{f}}(\vec{x}, \vec{u}, t),\,\, \vec{x}(t_s) = \vec{x}_s, \label{eq:oc-dynsys}
\end{align}
where $[t_s,t_f]$ is the time interval of interest;
$\vec{x} {:} [t_s, t_f] \mapsto \R^{n_x}$ is the state vector;
$\varphi : \R^{n_x} \mapsto \R$ is a terminal cost function;
$L { :} \R^{n_x} \times \R^{n_u} \times \R \mapsto \R$ is an intermediate cost function; and
$\vec{\tilde{f}} : \R^{n_x} \times \R^{n_u} \times \R \mapsto \R^{n_x}$ is a vector field.
Eq.~\eqref{eq:oc-dynsys} represents the dynamics of the system and its initial state. This problem definition is known as the Bolza problem; which for $\varphi(x(t_f)) = 0$ and $\vec{u} = \dot{\vec{x}}(t)$ it is known as the Lagrange problem \cite{goldstine2012history}. The cost function (or performance index) $\Gamma$ is a functional, which assigns a real value to each control function $\vec{u}$.

Often, the solution to  a control problems cannot be found by analytical means. Then, computational methods are used to solve such problems. Depending on the types of cost functions, time domain, and constraints in Eqs. (\ref{eq:oc-gamma})-(\ref{eq:oc-dynsys}) different methods may be applied, cf. Fig.~\ref{fig:optimal-control}. The direct methods use a discretization of the control problem and solve it using nonlinear programming approaches. Other methods involve the discretization of the differential equations by defining a grid of $N$ points covering the time interval $[t_s, t_f]$, $t_s = t_1 < t_2 < \ldots < t_N = t_f$, and solving these equations using suitable numerical methods \cite{press2007numerical}. Thereby, the differential equations become equality constraints of the nonlinear programming problem. Other direct methods involve the approximation of control and states using basis functions, such as splines or Lagrange polynomials.
%

Dynamic programming is an alternative to the variational approach to optimal control. It was proposed by Bellman in the 1950s and is an extension of Hamilton--Jacobi theory. A number of books exist on these topics including \cite{Lewis12995OptimalControl,Kirk1970OptimalControl,Bryson1975OptimalControl}.
A general overview of the optimal control methods for dynamical systems can be found in \cite{Becerra2008OptimalControl}. For further details, see~\cite{Athans2006OptimalControl}.

Our approach to solve a control problem uses machine learning to determine the optimal control. Therefore, we adopt the continuous-time formulation given in Eqs. \eqref{eq:oc-gamma} and \eqref{eq:oc-dynsys}. For multi-objective optimization, the reader is referred to~\cite{Gout2018}. In real-worl problems, often the derivatives are not given and one has to reconstruct them; then it is particularly important to respect the accuracy of measured data as described in~\cite{AHNERT2007764}. For the particular control scheme considered here, $\vec{\tilde{f}}$ and $\vec{u}$ are slightly reformulated, as will be described next.

\begin{figure}[htpb]
	\centering
	\includegraphics[width=.4\textwidth]{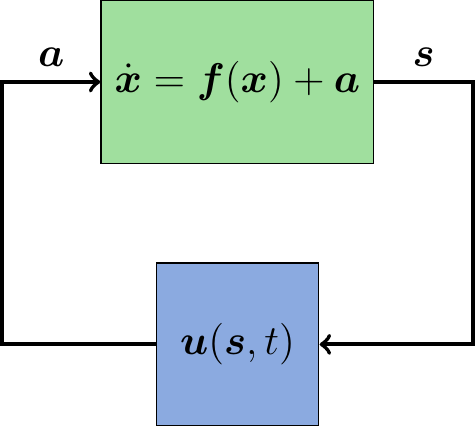}
	\caption{Sketch of the feedback control loop. The output of the dynamical system $\vec{x}$ is measured by sensors $\vec{s}$ which are used as input to the control function $\vec{u}$. The control function, in turn, acts on the system via actuators $\vec{a} = \vec{u}(\vec{s}, t)$ in order to achieve a desired state. (External disturbances which can be incorporated explicitly as additional inputs to the dynamical system and the control function are not shown here.) \label{fig:control-loop}}
\end{figure}

Similar to reinforcement learning, a feedback control scheme \cite{Gene2014FeedbackControl} is used here to implement the control. In Fig.~\ref{fig:control-loop} the architecture is depicted.
To follow this scheme, we rewrite Eq.~\eqref{eq:oc-dynsys}:
\begin{align*}
\dot{\vec{x}} = \vec{f}(\vec{x}, t) + \vec{a},\,\, \vec{x}(t_s) = \vec{x}_s,
\end{align*}
such that the uncontrolled system $\dot{{\vec{x}}} = \vec{f}(\vec{x}, t)$ is controlled by an additive actuator term $\vec{a}$, and the control function $\vec{u}$ depends on sensor measurements $\vec{s} \in \R^{n_s}$:
\begin{align*}
\vec{a} = \vec{u}(\vec{s}, t).
\end{align*}
These measurements might be nonlinear functions of the state $\vec{x}$. For simplicity, external perturbations to the dynamic system are not considered here.

\subsection{Genetic Programming \label{sec:gp}}

\begin{figure}[htpb]
	\centering
	\includegraphics[width=.3\textwidth]{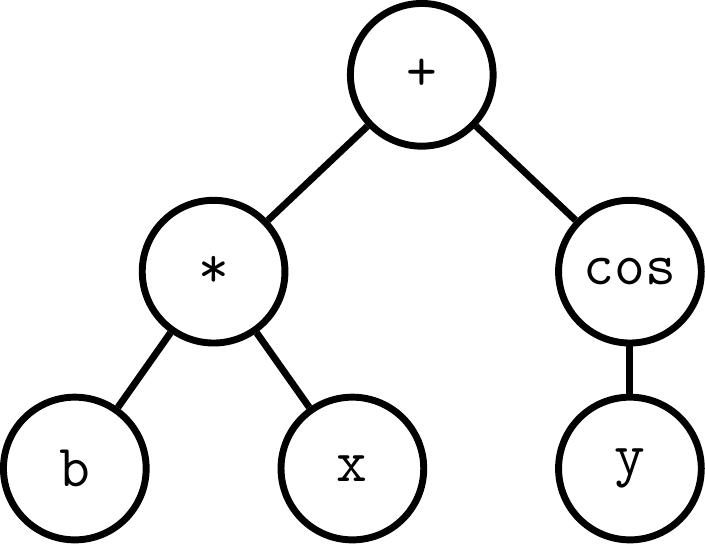}
	\caption{Tree representation of the mathematical expression $b \cdot x + \cos(y)$. The symbols \texttt{b}, \texttt{x}, and \texttt{y} are taken from the terminal set and make up the leaf nodes of the tree, whereas the symbols \texttt{*}, \texttt{+}, and \texttt{cos}, are symbols taken from the function set, they make up the internal nodes. \label{fig:gp_tree}}
\end{figure}

To obtain a solution for the control, we use the fairly general genetic programming (GP). This choice is motivated by its generality: in contrast to, e.g.,  generalized linear regression, no additive structure is needed as used already 20 years ago in~\cite{PhysRevLett.83.3422,PhysRevE_57_2820}.
GP \cite{Koza1992GeneticProgramming,Cramer1985Representation} is an evolutionary algorithm for global optimization. Similar to a genetic algorithm (GA), GP uses the natural selection metaphor to evolve a set of solutions using a cost-based selection mechanism. Often the bio-inspired terms \textit{population} and \textit{individual} are used correspondingly.
The evolution occurs over a number of iterations (\textit{generations}). GP differs from GA mainly in the representation of a solution: In GP, it is generally represented using lists or expression trees. Expression trees are constructed from the elements of two predefined primitive sets: a function set consisting of mathematical operators and functions, such as $\{$\texttt{+, -, *, cos, sin}$\}$, and a terminal set consisting of variables and constants, such as $\{$\texttt{ x, y, b}$\}$. Function symbols represent the internal nodes of a tree; and terminal symbols are used in the leaf nodes. For example, \ref{fig:gp_tree} shows the tree representation for the expression $b \cdot x + \cos(y)$. All elements of the tree are drawn from the aforementioned primitive sets: the variables and constants in the terminal set ($x$, $y$, and $b$) form the leaves of the tree and the mathematical symbols in the functional set ($\cdot$, $+$, and $\cos$) are used in forming the tree's internal nodes.

\begin{algorithm}[H]
	\begin{algorithmic}
		\Procedure{main}{}
		\State $G_0 \leftarrow \text{random}(\lambda)$
		\State $\text{evaluate}(G_0)$
		\State $t \leftarrow 1$
		\Repeat
		\State $O_t \leftarrow \text{breed}(G_{t-1}, \lambda)$
		\State $\text{evaluate}(O_t)$
		\State $G_{t} \leftarrow \text{select}(O_t, G_{t-1}, \mu)$
		\State $t \leftarrow t + 1$
		\Until{$t > T \textbf{ or } G_t = \text{good}()$}
		\EndProcedure
	\end{algorithmic}
	\caption{Top level description of a GP algorithm}
	\label{alg:mumu}
\end{algorithm}

\begin{figure}[htpb]
	\centering
	\includegraphics[width=.8\textwidth]{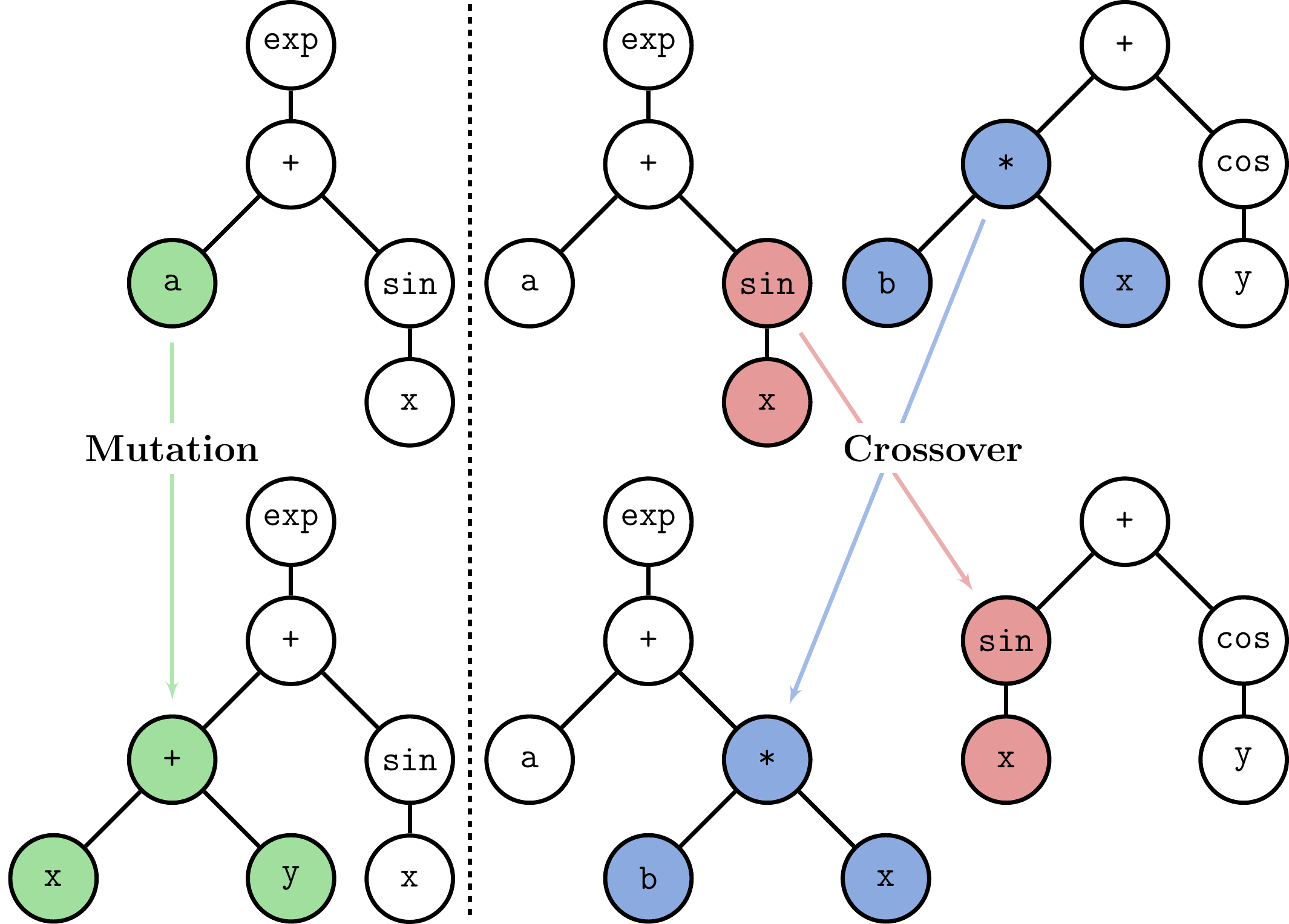}
	\caption{Breeding: Mutation and crossover operations on expression trees. \emph{Source:} Adapted from \cite{Quade2016Prediction}; used with permission. \label{fig:gp-mutation-and-crossover}}
\end{figure}

The GP algorithm is described below (~\ref{alg:mumu}): it starts with the initial generation of a population of random solutions $G_0$. A random solution is generated with a set maximum tree depth by choosing randomly operators, functions and variables. Each solution is then evaluated using the cost function that belongs to the problem. This cost is assigned to each solution, typically how closely a solution predicted the target function output. A new population of solutions $O_t$ is then generated by:
(i) probabilistic selection of parent solutions from the existing population using a cost-proportional selection mechanism, and
(ii) creation of offsprings by recombination (or crossover) and variation (or mutation) operators (see \ref{fig:gp-mutation-and-crossover}).
This procedure is repeated until the cost is reasonably low (the exact definition of low depends on the problem)  or a certain preset number of solutions (a fixed population size) is reached. The validity of generated solutions is ensured by a closure property, both for the initialization and breeding operations. Often, convergence is sped-up choosing a reproduction of the best $N$ solutions (elitist approach), then these best solutions are copied to the next-generation population $G_{t+1}$. The selection, evaluation and reproduction processes are repeated until one of the above criteria is met. For further details about GP operation, see~\cite{Poli2008FieldGuide,Yang2011Metaheuristics}.

\begin{figure}[htpb]
	\centering
	\includegraphics[width=.8\textwidth]{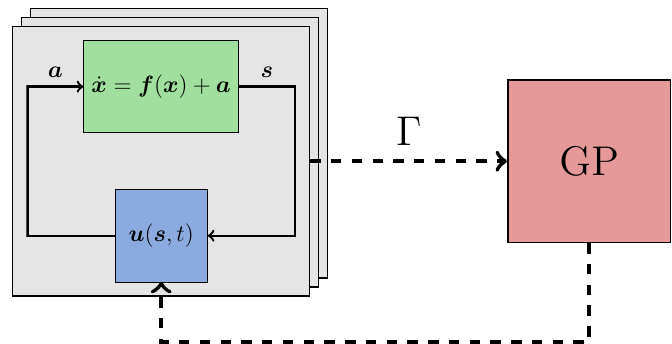}
	\caption{Sketch of the machine learning loop. By its evolutionary strategy, the GP algorithm generates a set of candidate control solutions $\vec{u}$, called the population. The candidate solutions are then evaluated in many realizations of the control loop; the performance in each iteration is rated via a cost functional $\Gamma$ and fed back as a cost index into the GP algorithm. The performance rating is used to select the best solutions and to evolve them into the next generation of candidate solutions. This learning loop repeats until at least one satisfactory control law is found (or other break conditions are met). \label{fig:learning-loop}}
\end{figure}

To solve a general control problem with GP, it is formulated as a learning and optimization task. That is, we learn a control function using GP which drives and keeps a dynamical system in a desired state. The typical choice for the cost function ($\vec{\Gamma}$) is the difference between a given state in time and the desired state. 
This function $\vec{\Gamma}$ can possess complex properties, like non-linearity, multi-modality, multi-variability and discontinuity. Many traditional direct and gradient methods can not handle such properties, however, meta-heuristic methods, such as GP, are suitable candidates  for this task. In Fig.~\ref{fig:learning-loop} a GP-based dynamic controller within a feedback control loop is sketched, shown in \ref{fig:control-loop}.

The treatment of multiobjectivity and constant optimization is explained in \cite{Gout2018} and will not be touched here. Rather we focus on the analysis of the analytical expressions resulting from our control optimization. To reproduce the results shown below, we have given the concrete setup for GP used here in the appendix~\ref{sec:appendix}.

\section{Results
	\label{sec:results}
}

In this section, we explain the concrete application we use to illustrate the power of our explainable GP: in a previous publication, GP-based control has been used for the control of networks of oscillators \cite{Gout2018}. Such networks are used to model highly nonlinear complex systems, including the human brain. For our purposes, we systematically investigate the results of our method starting using two coupled oscillators. The extension to many oscillators is straightforward and subject of ongoing implementation activities to include a stability analysis automatically into \textit{Glyph}~\cite{glyph}.

The aim, of our consideration is to control the synchronization behavior of the coupled oscillators. This can be done in two ways: starting from a synchronization regime and forcing the system into de-synchronization or vice versa, i.e., starting from a de-synchronized regime and forcing the system into synchronization. Both control goals are evaluated in~\cite{Gout2018}. Here, we focus on synchronization control, since we mainly want to demonstrate the power of symbolic regression methods as explainable, rigorously treatable models and control terms, respectively.

\begin{figure}
	\centering
	\includegraphics[width=1.0\textwidth]{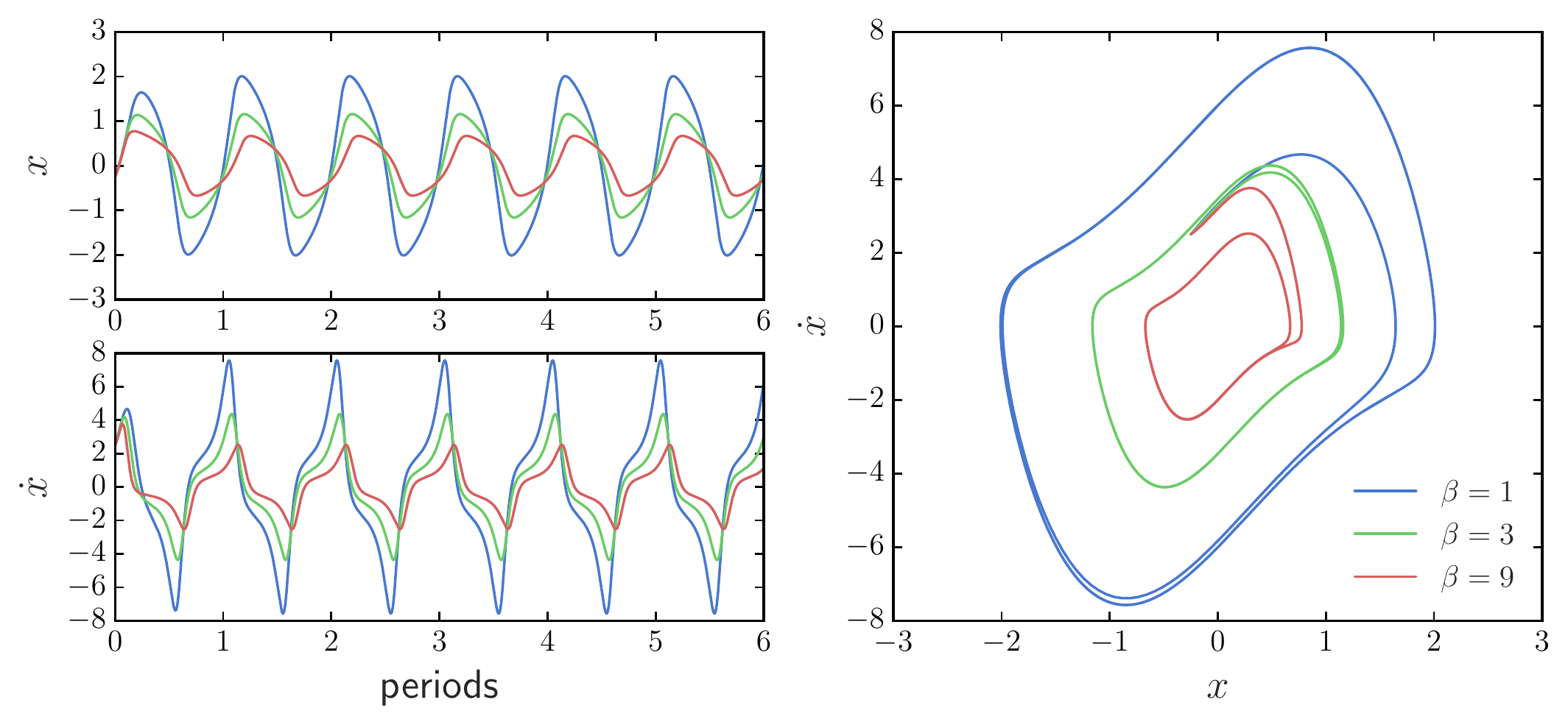}
	\caption{Single van der Pol oscillator (with parameters $\omega=e^2$, $\alpha=3$, and initial conditions $x(0) = -0.25$, $\dot{x}(0) = 2.5$.)\label{fig:van_der_pol}}
\end{figure}

Let us first and briefly discuss synchronization again.
The synchronization of dynamical systems is well-known exhibited by a huge variety of oscillators and oscillatory media \cite{Pikovsky2003Synchronization}. Here, we use a popular  model, the van der Pol oscillator, also used as a simple model for neurons:
\begin{align}
\ddot{x} = - \omega^2 x + \alpha \dot{x} \left( 1 - \beta x^2 \right) =: f_{\text{vdP}}(x, \dot{x}),\label{eqn:van-der-pol}
\end{align}
where $x$ is the state and $\omega$, $\alpha$, $\beta > 0$ are model parameters. The parameter $\omega$ is the frequency at which the system oscillates without any driving or damping force. The parameter $\alpha$ controls the non-linearity of the system: if $\alpha=0$, Eq.~\eqref{eqn:van-der-pol} is a harmonic oscillator equation. The damping parameter $\beta$ controls the nonlinear deformation of the trajectory in phase space. See \figref{fig:van_der_pol}.
In our setup, we reproduce the results of~\cite{Gout2018} and use linear coupling. An uncontrolled system of $N$ coupled van der Pol oscillators can be stated as follows:
\begin{align}
\ddot{x}_i = f_{\text{vdP}}(x_i, \dot{x}_i) + c_1 \sum_{j=0}^{N-1} \kappa_{ij} x_j + c_2 \sum_{j=0}^{N-1} \varepsilon_{ij} \dot{x}_j \qquad (i = 0,\ldots,N-1)\label{eqn:coupled-van-der-pol}
\end{align}
with initial conditions
\begin{align*}
x_{i}(t_0) = x_{i,0},\qquad \dot{x}_{i}(t_0) = \dot{x}_{i,0},
\end{align*}
where $c_{1,2}$ are the global coupling constants and $(\kappa_{ij})$ and $(\varepsilon_{ij})$ are the respective coupling matrices. This allows for several types of coupling such as direct, diffusive, and global coupling, or any other kind of network-like coupling. In the following experiments, we will use diffusive coupling in $\dot{x}_i$. For GP, we use the same setup described above (\secref{sec:appendix}).


\subsection{Two Coupled Oscillators\label{sec:two_vdp}}

For our further investigations we use the simplest system showing synchronization: two
diffusively coupled van der Pol oscillators:
\begin{align}
\begin{split}
\ddot{x}_0 = f_{\text{vdP}}(x_0, \dot{x}_0) + c \left( \dot{x}_1 -
\dot{x}_0 \right),\\ \label{eqn:paired_vdp}
\ddot{x}_1 = f_{\text{vdP}}(x_1, \dot{x}_1) + c \left( \dot{x}_0 -
\dot{x}_1 \right).
\end{split}
\end{align}
The coupling is restricted to $\dot{x}_i$, in which case the coupling
constants from \eqref{eqn:coupled-van-der-pol} are set to $c_1 = 0$,
$c_2 = c$, and the remaining coupling matrix reads $
(\varepsilon_{ij}) =
\begin{bmatrix}
-1 &  1 \\
1 & -1
\end{bmatrix}
$.
In the case of zero coupling, $c=0$, some parameter combinations $(\alpha, \beta)$ allow stable limit cycles with characteristic frequencies $\omega_{0,1}$. If the coupling constant $c \neq 0$, a range of frequencies with $\omega_0 \neq \omega_1$ exists, where both oscillators oscillate with exactly the same frequency $\Omega$ in a common mode. This range of frequencies is called the synchronization region. With variation of the coupling constant this region changes in width.

For an illustration, we chose quite arbitrary $\alpha=0.1$, $\beta=1$, with $\omega_0 = 1.386$. The harmonic frequency $\omega_1$ of the second oscillator is varied  in the range $[\omega_0  - 0.06,\, \omega_0 + 0.06]$. By the above explanation, one expects a range where both oscillators have a common, observed frequency  $\Omega_0=\Omega_1=\Omega$, such that $\Delta \Omega=\Omega_1-\Omega_0=0$, and a range with $\Delta \Omega \neq 0$. This frequency $\Omega$ is determined numerically by Fourier transform.

\begin{figure}
	\centering
	\includegraphics[width=0.8\textwidth]{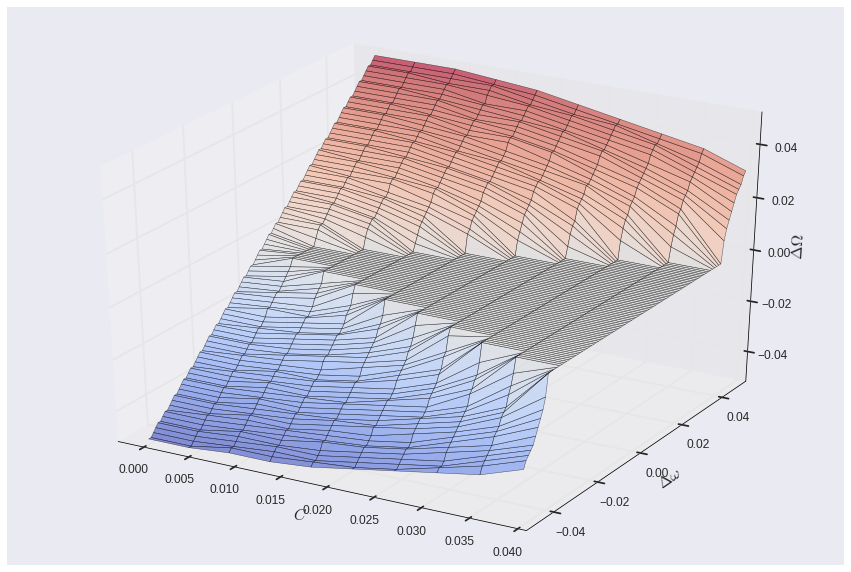}
	\caption[Synchronization of two coupled van der Pol oscillators]{Synchronization plot of two coupled van der Pol oscillators with varying coupling strength $c$. The horizontal V-shaped plateau is referred to as the Arnold tongue, it represents regions of synchronization. (The parameter set and initial conditions used are stated in the left part of \tabref{tab:two_vdp_sync_setup}.)}
	\label{fig:van_der_pol_sync_plot_full}
\end{figure}

For a visualization, $\Delta \Omega$, is plotted  against the difference in their characteristic frequencies, $\Delta\omega := \omega_1 - \omega_0$, cf.  \figref{fig:van_der_pol_sync_plot_full}. Regions of synchronization show up as horizontal segments at $\Delta\Omega = 0$ (also, note the symmetry about $\Delta \omega = 0$). If we do this for several values $c$ in the range $[0, 0.4]$ we can trace out the regions of synchronization: The result is a typical V-shaped plateau, the Arnold tongue.
Reading off suitable parameters from Fig.~\figref{fig:van_der_pol_sync_plot_full} allows to choose appropriate parameters $\omega_1$ and $c$ to setup the system for control, such that, in its uncontrolled state, it follows either the synchronization regime or the de-synchronization regime. This same approach is taken for all the experiments presented in this section, but will not be explicitly stated beyond this point. We show results only for a control of the state into synchronization, the desynchronization way works very similar, however the tracing of the stability of the solution becomes more tedious, because the solution is generically quasiperiodic and no longer periodic (if the two frequencies $\omega_0$, $\omega_1$ are incommensurate).

Let us add the control function, $u$, to the equations \eqref{eqn:paired_vdp} of the uncontrolled system:
\begin{align}
\begin{split}
\ddot{x}_0 = f_{\text{vdP}}(x_0, \dot{x}_0) + c \left( \dot{x}_1 - \dot{x}_0 \right) + u(\dot{\vec{x}}),\\
\ddot{x}_1 = f_{\text{vdP}}(x_1, \dot{x}_1) + c \left( \dot{x}_0 - \dot{x}_1 \right) + u(\dot{\vec{x}}).
\label{eqn:paired_vdp_controlled}
\end{split}
\end{align}
The actuation $u$ may depend on $\dot{x}_0$ and $\dot{x}_1$, summarized in vector notation as $\dot{\vec{x}} = (\dot{x}_0, \dot{x}_1)$; it  is added as a global actuator term with equal influence on both oscillators, this role may be changed into more complex scenarios.

\subsubsection{Forced Synchronization}

The system setup for forced synchronization of the two coupled van der Pol oscillators is presented in the appendix, Tab.\tabref{tab:two_vdp_sync_setup}. The parameters $\omega_1$ and $c$ are chosen according to \figref{fig:van_der_pol_sync_plot_full}, such that the uncontrolled system follows a de-synchronization regime at a distance, $\Delta \omega$, approximately half the plateau from the closest synchronization point. The initial conditions are the same for both oscillators. 

The degree of de-synchronization is encompassed by the cost functional
\begin{align}
\begin{split}
\Gamma_1 := |\Omega_0 - \Omega_1|.\label{eqn:two_vdp_sync_cost}
\end{split}
\end{align}
It measures the difference in observed frequencies exhibited by the two oscillators: smaller differences reduce the cost on this objective.

As stated in the previous section, the actual frequencies, $\Omega_0$ and $\Omega_1$, are numerically determined by counting zero crossings of the trajectory $x - \langle x \rangle$. This requires a careful choice of the time range $[t_0, t_n]$ of observation, since the number of periods, $N_P$, fitting into this interval determines an upper bound in absolute accuracy ($\sim\tfrac{1}{2N_P}$) of measuring $\Omega_0$, $\Omega_1$. Here, $N_P = 2000$ to yield an absolute accuracy well below $10^{-3}$ in the frequency range of interest.

\begin{figure}[htpb]
	\centering
	\includegraphics[width=1.\textwidth]{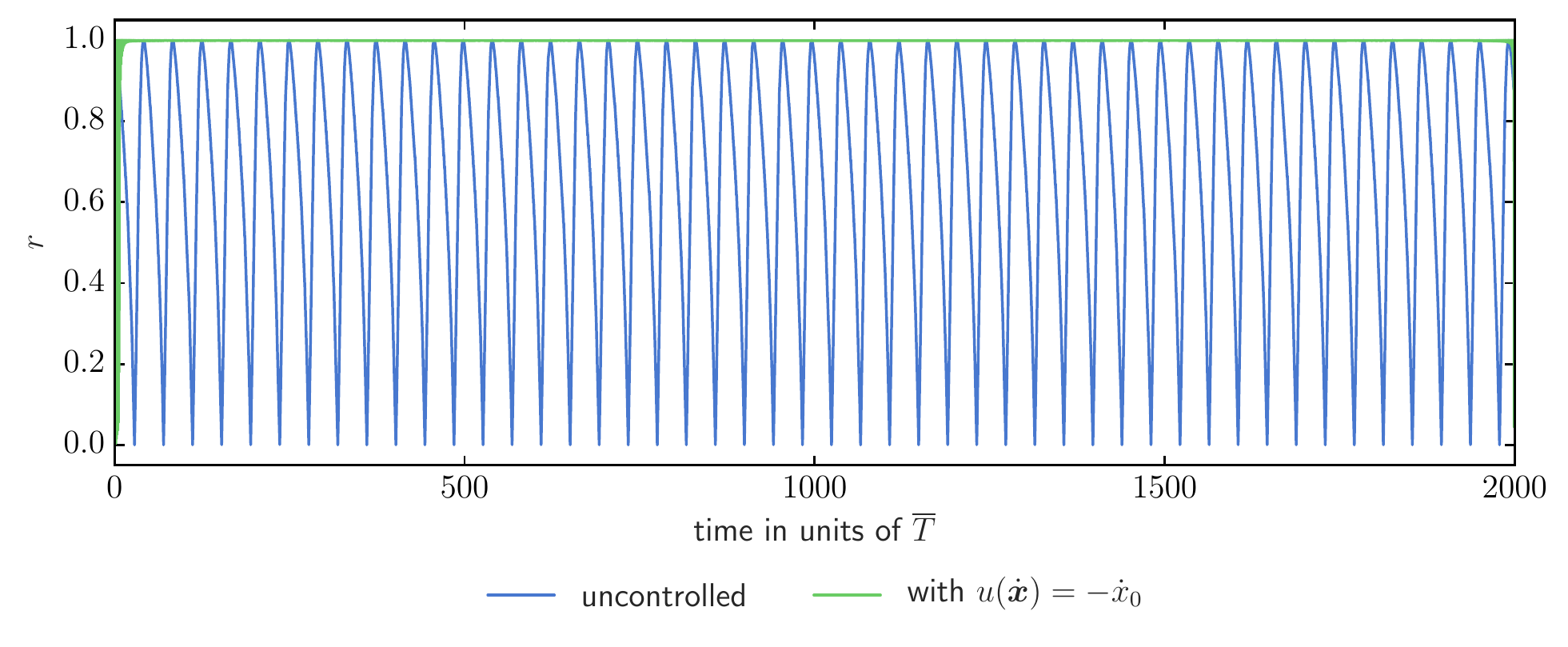}
	\caption[Two coupled oscillators sync: kuramoto]{Two Coupled Oscillators: Kuramoto order parameter, $r$, for forced synchronization. Green: the controlled, and blue: the uncontrolled system.\label{fig:two_vdp_sync_kuramoto}}
\end{figure}

The top six control laws found are given in Tab.~\tabref{tab:two_vdp_sync_results}. The algorithm stopped after one generation, providing six simple results optimally satisfying $\Gamma_1$, the synchronization criterion. At this point we can start already to \textit{explain} the results. First, one notes that for each term in $x_0$ we find a counterpart in $x_1$. This can be explained by the symmetry in Eqs.~\eqref{eqn:paired_vdp_controlled}. But why, then, are the control terms not symmetric themselves, in the frequencies? This is at first sight not logical, however, if we check our cost functional, we recognize that we only enforce synchronization, and not symmetry of the solution. In particular, we note that the amplitude of one oscillator might vanish while the other one is controlling it. That way, the result makes sense. A deep learning result would not allow immediately such a simple and clear insight.
\begin{table}[htpb]
	\caption[Two coupled oscillators sync: solutions]{Two Coupled Oscillators: Optimal solutions for forced synchronization.\label{tab:two_vdp_sync_results}}
	\centering
	\begin{tabular}{*{4}{c}}
		\toprule
		$|\Omega_0 - \Omega_1|$ & length & \mc{expression} & \\
		\midrule
		\addlinespace
		$0.0$ & $ 2$ & $\cos(\dot{x}_1)$ &  \\
		$0.0$ & $ 2$ & $\cos(\dot{x}_0)$ &  \\
		$0.0$ & $ 2$ & $-\dot{x}_0     $ &  \\
		$0.0$ & $ 2$ & $\sin(\dot{x}_1)$ &  \\
		$0.0$ & $ 2$ & $-\dot{x}_1     $ &  \\
		$0.0$ & $ 2$ & $\sin(\dot{x}_0)$ &  \\
		\addlinespace
		\bottomrule
	\end{tabular}
\end{table}
To demonstrate the control effect \figref{fig:two_vdp_sync_kuramoto} shows the Kuramoto order parameter, $r$, for the particular solution $u(\dot{\vec{x}}) = -\dot{x}_0$. The parameter represents phase-coherence over time \cite{Kuramoto1975,Kuramoto1984} and is defined as
\begin{align*}
r &= \left| \frac{1}{N} \sum_{j=0}^{N-1} e^{i\varphi_j} \right|,
\end{align*}
with $\varphi_j$ being the continuous phase of the $j$-th oscillator. This continuous phase is computed from the analytic signal of the trajectory using the Hilbert transform, cf. \cite{Cohen1994TimeFrequencyAnalysis}. The plot shows, that the controlled system completely synchronizes ($r \approx$ const.) after passing through a short initial period of de-synchronization; whereas the uncontrolled system exhibits a permanent phase shift resulting in an oscillating graph.
Now it is of highest interest to understand how robust the found control law is against perturbation. This is a serious study one has to do. Using explainable MLC we can use the well understood and rigorous mathematical framework of stability analysis. In particular for nonlinear and complex systems that leads us onto safe ground whereas a neural network solution would require more, mor complicated and very expensive studies. 
Before going into detail with stability 
in Sec.~\secref{sec:bifurcation-analysis} we reiterate results for forced de-synchronization.

\subsubsection{Forced De-Synchronization}

The system setup for forced de-synchronization is given in the appendix in Tab.~\tabref{tab:two_vdp_sync_setup}. The parameters $\omega_1$ and $c$ are, again, chosen according to \figref{fig:van_der_pol_sync_plot_full}. This time, such that the uncontrolled system follows a synchronization regime well inside the plateau. The measure for the degree of synchronization is now reciprocal to the previous case
\begin{align}
\Gamma_1 := \exp(-|\Omega_0 - \Omega_1|),
\label{eqn:two_vdp_desync_cost}
\end{align}
and penalizes synchronization of the two oscillators. Other GP parameters are the same as for forced synchronization.

\begin{figure}[htpb]
	\centering
	\includegraphics[width=1.\textwidth]{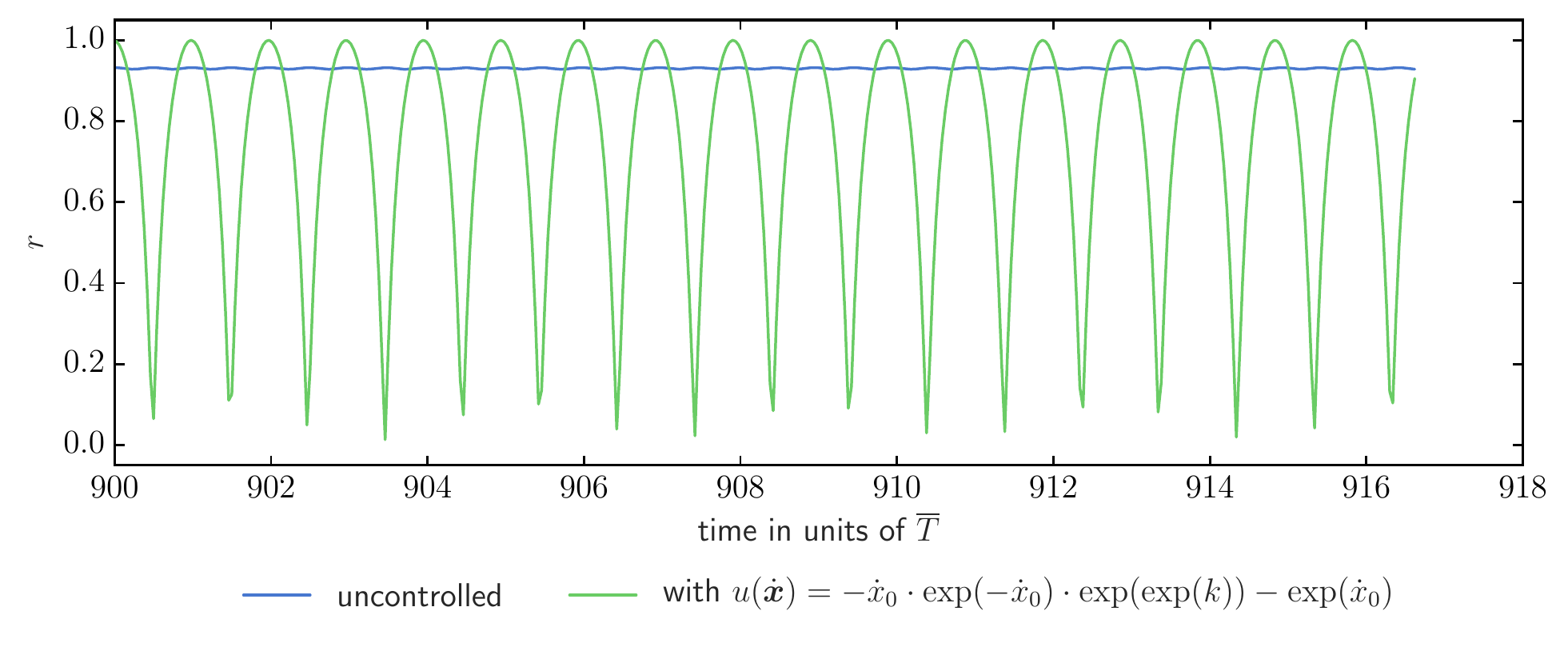}
	\caption[Two coupled oscillators de-sync: kuramoto]{Two Coupled Oscillators: Kuramoto order parameter, $r$, for forced synchronization. Green: the controlled, and blue: the uncontrolled system. The horizontal axis is scaled to a limited time window in order to make the oscillations visible.}
	\label{fig:two_vdp_desync_kuramoto}
\end{figure}

In Tab.~\tabref{tab:two_vdp_desync_results}, we show results from the GP run. These results are, again, exact reproductions of \cite{Gout2018}. Interestingly, of the 8 solutions found, only the best two are worth being called desynchronized. It indicates that it is much harder to synchronize a desynchronized solution than vice versa. Then, the control term is a long expression in contrast to the ones found for synchronization. As a further fact, constant optimization seems to fail in all cases where a constant is present (this is expressed by a value $k=1$, which corresponds to the initial guess of the optimization procedure). Still, the oscillating Kuramoto parameter, $r$, of the controlled system in \figref{fig:two_vdp_desync_kuramoto} shows, that the best solution with respect to $\Gamma_1$ performs well in de-synchronizing the oscillators. 

Now let us interpret this solution. A look shows $u(\dot{\vec{x}}) = -\dot{x}_0 \cdot \exp(\exp(k) + \cos(k)) = - \tilde{k} \dot{x}_0$, with $\tilde{k}\approx 26$. This term has the same structure as one of the best solutions found to enforce synchronization, namely the control law $u(\dot{\vec{x}}) = -k\dot{x}_0$, with coefficient $k=1$. Both are analyzed in more detail in \secref{sec:bifurcation-analysis}.

\begin{table}[htpb]
	\caption[Two coupled oscillators de-sync: solutions]{Two Coupled Oscillators: Pareto-front solutions for forced
		de-synchronization.\label{tab:two_vdp_desync_results}}
	\centering
	\begin{tabular}{ccll}
		\toprule
		$\exp(-|\Omega_0 - \Omega_1|)$ & length & \mc{expression} & \mc{constant} \\
		\midrule
		\addlinespace
		$0.248$ & $ 9$ & $-\dot{x}_0 \cdot \exp(\exp(k) + \cos(k))     $ & $k = 1$ \\
		$0.258$ & $ 7$ & $\cos(\exp(\dot{x}_1 + \cos(\cos(\dot{x}_0))))$ &  \\
		$0.875$ & $ 4$ & $\cos(\exp(\exp(\dot{x}_0)))                  $ &  \\
		$0.912$ & $ 3$ & $\sin(\exp(\dot{x}_0))                        $ &  \\
		$0.999$ & $ 2$ & $\exp(k)                                      $ & $k = 1$ \\
		$1.000$ & $ 1$ & $\dot{x}_1                                    $ &  \\
		$1.000$ & $ 1$ & $\dot{x}_0                                    $ &  \\
		$1.000$ & $ 1$ & $k                                            $ & $k = 1$ \\
		\addlinespace
		\bottomrule
	\end{tabular}
\end{table}

The simplification $u(\dot{\vec{x}}) = -\dot{x}_0 \cdot \exp(\exp(k) + \cos(k))$ to $u(\dot{\vec{x}}) = -26 \cdot \dot{x}_0$, suggests the question, why the GP algorithm did not directly generate this simpler solution. One reason is the stop criterion which prevents solutions to converge further to simplified version. On the other hand, on failure, the least squares algorithm returns the result of the last internal iteration. This return value might be entirely inadequate for $k$, which, in turn, could lead to an large cost $\Gamma_1$, and by further integration of  the dynamic system \eqref{eqn:paired_vdp_controlled} the corresponding solution is discarded.

\subsubsection{Control Terms and Bifurcation Analysis\label{sec:bifurcation-analysis}}

The objective of this section is to analyze the effect of control exhibited by the particular results $u(\dot{\vec{x}}) = -k \cdot \dot{x}_0$ found on the synchronization (and desynchronization) of the two oscillators.
To study synchronization, we first note that it is adequate to reduce Eqs.~\ref{eqn:paired_vdp} by one dimension. This is done by splitting off the fast oscillation and averaging methods~\cite{Pikovsky2003Synchronization}. The result is a system of three first order ODEs which can be studied with respect to stability. We use the well-known package \cite{AUTO07p} to track branches of their solutions. 
This would not be possible at all for a neural network control due to the complex network structure and high dimension.

Let us first dwell on the analytical work we can do now.
One plugs $u(\dot{\vec{x}}) = -k \dot{x}_0$ into the first oscillator equation from \eqref{eqn:paired_vdp_controlled}:
\begin{align}
\begin{split}
\label{eqn:paired_vdp_result_0}
\ddot{x}_0 &= - \omega_0^2 x_0 + \alpha \dot{x}_0 \left( 1 - \beta x_0^2 \right) + c \left( \dot{x}_1 - \dot{x}_0 \right) - k \dot{x}_0\\
&= - \omega_0^2 x_0 + (\alpha - c - k)\dot{x}_0 - \alpha \beta \dot{x}_0 x_0^2 + c \dot{x}_1 \\
&= - \omega_0^2 x_0 + a_0 \dot{x}_0 - b \dot{x}_0 x_0^2 + c_0 \dot{x}_1,
\end{split}
\end{align}
with $a_0 := \alpha - c - k$, $b := \alpha \beta$, and $c_0 := c$. For the second oscillator equation one obtains
\begin{align}
\begin{split}
\label{eqn:paired_vdp_result_1}
\ddot{x}_1 &= - \omega_1^2 x_1 + \alpha \dot{x}_1 \left( 1 - \beta x_1^2 \right) + c \left( \dot{x}_0 - \dot{x}_1 \right) - k \dot{x}_0\\
&= - \omega_1^2 x_1 + (\alpha - c) \dot{x}_1 - \alpha \beta \dot{x}_1 x_1^2 + (c - k)\dot{x}_0\\
&= - \omega_1^2 x_1 + a_1 \dot{x}_1 - b \dot{x}_1 x_1^2 + c_1 \dot{x}_0,
\end{split}
\end{align}
where $a_1 := \alpha - c$ and $c_1 := c - k$. This way, one obtains another system of two coupled van der Pol oscillators, however, with direct coupling in place.

As indicated, this system can be analyzed in the framework of synchronization theory which uses the method of averaging \cite{Pikovsky2003Synchronization} under the assumption that the system is weakly nonlinear. We repeat the calculations here for our system.
In essence, the second-order equations are rewritten as two first-order equations $\dot{x}_j = y$, $\dot{y}_j =$ r.h.s.  ($j=0,1$), where r.h.s. denotes the right hand side of \eqref{eqn:paired_vdp_result_0} and \eqref{eqn:paired_vdp_result_1} respectively.
Next, the transformation
\begin{align}
x_j &= \frac{1}{2}(A_j e^{i\omega t} + A_j^* e^{-i\omega t}),\\
y_j &= \frac{1}{2}(i\omega A_j e^{i\omega t} - i\omega A_j^* e^{-i\omega t}),
\end{align}
is applied, where $j=0,1$, and $A_j(t) = R_j(t) e^{i \Theta_j(t)}$ is the time-dependent complex amplitude. This ansatz yields differential equations for the real amplitude $R$ and the phase $\Theta$, which are both slowly varying. They result from the collection of terms with vanishing fast oscillation. 

When one writes the equations \eqref{eqn:paired_vdp_result_0} and \eqref{eqn:paired_vdp_result_1} as 
\begin{align}
\dot{y}_0 &=  - \omega_0^2 x_0 + a_0 y_0 - b y_0 x_0^2 + c_0 y_1,\\
\dot{y}_1 &=  - \omega_1^2 x_1 + a_1 y_1 - b y_1 x_1^2 + c_1 y_0,
\end{align}
with the corresponding approximations (first order nonlinearities, slow dynamics) one arrives at new equations for $A$
\begin{align}
\dot{A}_0 &=  - i\Delta_0 A_0 + \frac{a_0}{2} A_0 - \frac{b}{8} |A_0|^2 A_0 + c_0 A_1,\\
\dot{A}_1 &=  - i\Delta_1 A_1 + \frac{a_1}{2} A_1 - \frac{b}{8} |A_1|^2 A_1 + c_1 A_0,
\end{align}
with $\Delta_j = \omega_j -\omega$ ($j = 0, 1$). For the real phases and amplitudes of $A(t)$ one then obtains a system of four real equations
\begin{align}
\begin{split}
\label{eq:amplitude_and_phase}
\dot{R}_0 &= \frac{a_0}{2} R_0 - \frac{b}{8} |R_0|^2 R_0  + c_0 R_1 \cos(\Theta_1 - \Theta_0),\\
\dot{R}_1 &= \frac{a_1}{2} R_1 - \frac{b}{8} |R_1|^2 R_1  + c_1 R_0 \cos(\Theta_0 - \Theta_1),\\
\dot{\Theta}_0 &= - \Delta_0 + c_0 \frac{R_1}{R_0} \sin(\Theta_1 - \Theta_0),\\
\dot{\Theta}_1 &= - \Delta_1 + c_1 \frac{R_0}{R_1} \sin(\Theta_0 - \Theta_1).
\end{split}
\end{align}
For the phase difference $\Theta_1 - \Theta_0$ an asymmetric so-called Adler-type equation results, which may or may not show synchronization for the given parameters:
\begin{align}
\Delta \dot{\Theta} = - \Delta \omega - (c_0 \frac{R_1}{R_0} + c_1\frac{R_0}{R_1}) \sin(\Theta_1-\Theta_0). \label{eq:Adler}
\end{align}
with $\Delta \omega := \omega_1 -\omega_0$.
If the stationary state for the phase difference $\Delta \dot{\Theta}=0$ has a solution, one can find synchronization, else not.
A special solution where also the amplitudes are stationary ($\dot{R}_0 = \dot{R}_1 =0$) has to be determined numerically. 

Thus, the full coupled system for stationary solutions of \eqref{eq:amplitude_and_phase} reads:
\begin{align}
\begin{split}
0 &= \frac{a_0}{2} R_0 - \frac{b}{8} |R_0|^2 R_0  + c_0 R_1 \cos(\Delta \Theta), \\
0 &= \frac{a_1}{2} R_1 - \frac{b}{8} |R_1|^2 R_1  + c_1 R_0 \cos(\Delta \Theta), \\
0 &= - \Delta \omega  + ( c_0 \frac{R_1}{R_0} - c_1 \frac{R_0}{R_1} ) \sin(\Delta \Theta),
\label{eq:controlled}
\end{split}
\end{align}
with parameters $a_0 = \alpha - c - k$, $a_1 = \alpha - c$, $b_0 = \alpha \cdot \beta$, $b_1 = \alpha \cdot \beta$, $c_0 = c$, $c_1 = c - k$ and $\Delta \Theta = \Theta_0 - \Theta_1$.

First one can observe, that the term $k\dot{x}_i$ ($i = 0, 1$) introduces an asymmetry in the equations \eqref{eq:amplitude_and_phase}, such that one or the other oscillator might be ``favored'' by the dynamics, since it may have a different damping depending on the parameter settings.
In an uncoupled system ($c=0$) this is of course relevant if one needs to study a real application.
For a better understanding of the implications of the control term $k\dot{x}_1$ we note that in the uncoupled case, the first equation of \eqref{eq:controlled} reduces to
\begin{align}
0 &= \frac{a}{2} R_0 - \frac{b}{8} |R_0|^2 R_0,
\label{eq:controlled-single}
\end{align}
with rescaled parameters $a = \alpha - k$, $b = \alpha \cdot \beta$.
This is the normal form of a pitchfork bifurcation.
In the original, non-averaged system, this corresponds to a Hopf-bifurcation because the full system shows oscillations.

To this end, we use the path-following and bifurcation analysis package \emph{AUTO-07p} \cite{AUTO07p}.
In the scope of this work, this analysis has been done manually, it is however straightforward to extend the MLC software to do this in an automated fashion for \textit{any} control term found.

\begin{figure}[htpb]
	\centering
	\hfill 
	\begin{overpic}[width=0.4\textwidth]{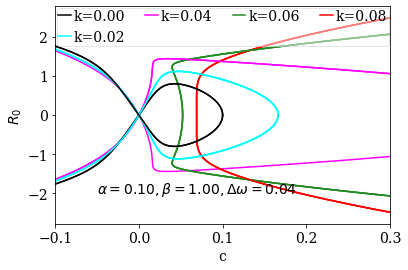} \put(-5,55){\textbf{(a)}} \end{overpic}\hfill
	\begin{overpic}[width=0.4\textwidth]{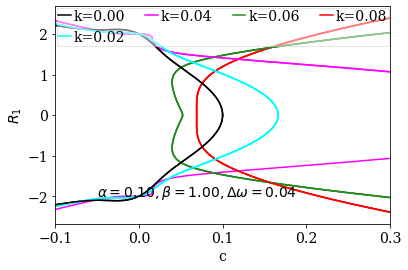} \put(-5,55){\textbf{(b)}} \end{overpic}
	\begin{overpic}[width=0.4\textwidth]{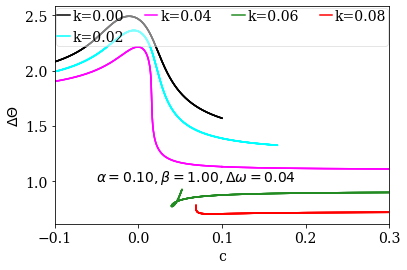} \put(-5,55){\textbf{(c)}} \end{overpic}
	
	\caption[Stationary solutions for the coupled system, varying $c$]
	{
		Stationary solutions for the system \eqref{eq:controlled}.
		Solution variables \textbf{a} $R_0$, \textbf{b} $R_1$ and \textbf{c} $\Delta\Theta$ are plotted against varying coupling $c$ and a set of fixed control parameters $k$.
		The trivial solution ($R_0=R_1=\Delta\Theta=0$) has been omitted in the plot for better overview.
	}
	\label{fig:bifurcation_vary_c}
\end{figure}

\begin{figure}[htpb]
	\centering
	\hfill
	\begin{overpic}[width=.4\textwidth]{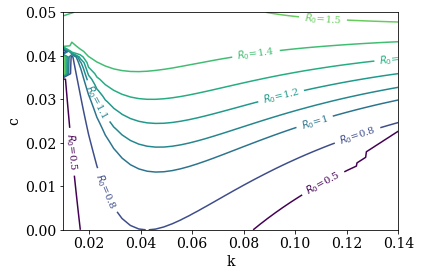} \put(-2,55){\textbf{(a)}} \end{overpic}
	\hfill 
	\begin{overpic}[width=.4\textwidth]{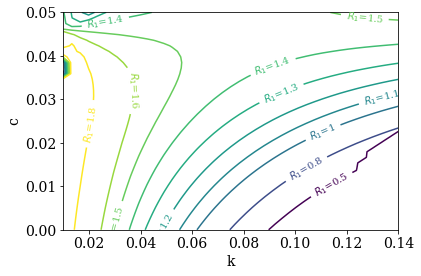} \put(-2,55){\textbf{(b)}} \end{overpic}
	\hfill
	
	\begin{overpic}[width=.4\textwidth]{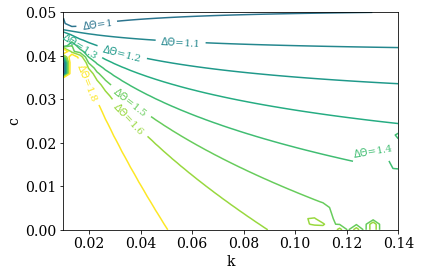} \put(-2,55){\textbf{(c)}} \end{overpic}
	\caption[Bifurcation in the parameter plane]{The bifurcation scenario in the parameter plane. The  contour lines correspond to  lines  of equal height of the solutions.  }
	\label{fig:bifurcation_2D}
\end{figure}

The paths of stationary solutions, as observed in the following, are interpreted by comparing to the standard textbook examples, as can be found in e.g. \cite{Strogatz2014NLD}.
First, we consider the uncoupled case ($c=0$) for varying $k$.
Next, solutions for $R_0, R_1, \Delta\Theta$ are tracked against varying $k$, or $c$, respectively, for several values of $c$, or $k$, respectively, cf.Fig.~\figref{fig:bifurcation_vary_k} and Fig.~\figref{fig:bifurcation_vary_c}, respectively.
Finally, contour lines for fixed component values of the stationary solution to \eqref{eq:controlled} are shown demonstrating that a wide variety of choices of particular synchronization details is possible by tuning the $k$ and $c$ parameters to the right values (\figref{fig:bifurcation_2D}).

\begin{figure}[htpb]
	\centering
	\begin{overpic}[width=.4\textwidth]{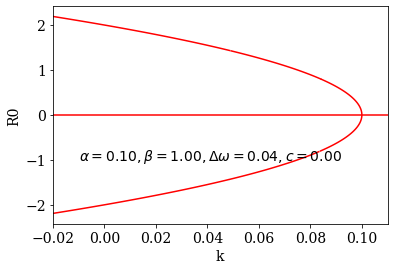} \put(-5,50){\textbf{(a)}} \end{overpic}
	\begin{overpic}[width=.4\textwidth]{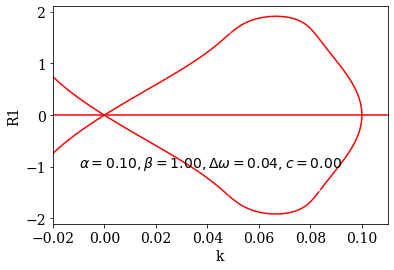} \put(-5,50){\textbf{(b)}} \end{overpic}
	\begin{overpic}[width=.4\textwidth]{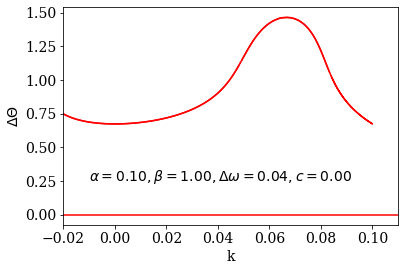} \put(-5,50){\textbf{(c)}} \end{overpic}
	\caption[Stationary solutions for the uncoupled system]
	{
		Stationary solutions for the system \eqref{eq:controlled} without coupling ($c=0$).
		\textbf{(a)} $R_0$,  \textbf{(b)} $R_1$, \textbf{(c)} $\Delta \Theta$ vs $k$ are plotted.
		Here, both trivial ($R_0=R_1=\Delta \Theta=0$) and non-trivial solution are shown.
		Beyond $k=\alpha$ no nontrivial solution is possible.
		Physically this reflects the fact that the damping is  so large  that no oscillation is  possible.
	}
	\label{fig:bifurcation_uncoupled}
\end{figure}

\begin{figure}[htpb]
	\centering
	\hfill 
	\begin{overpic}[width=0.4\textwidth]{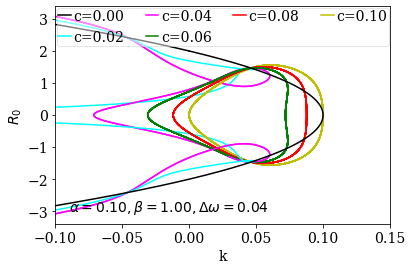} \put(-5,55){\textbf{(a)}} \end{overpic}
	\hfill 
	\begin{overpic}[width=0.4\textwidth]{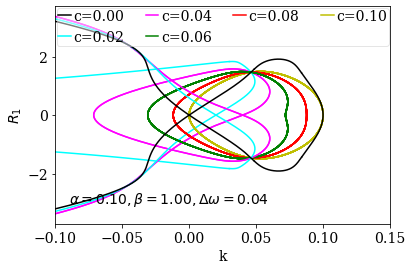} \put(-5,55){\textbf{(b)}} \end{overpic}
	\hfill
	
	\begin{overpic}[width=0.4\textwidth]{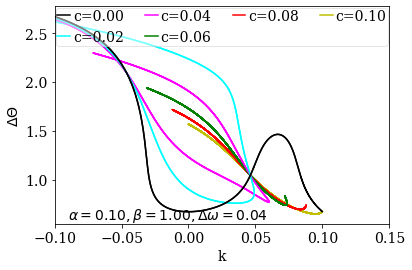} \put(-5,55){\textbf{(c)}} \end{overpic}
	\caption[Stationary solutions for the coupled system, varying $k$]
	{
		Stationary solutions for the coupled system \eqref{eq:controlled}.
		Solution variables \textbf{a} $R_0$, \textbf{b} $R_1$ and \textbf{c} $\Delta\Theta$ are plotted against varying control $k$ and a set of fixed couplings $c$.
		For $c=0$ the scenario of \figref{fig:bifurcation_uncoupled} is recovered (black line).
		With increasing coupling, the control term becomes relatively weaker and eventually coupling dominates the dynamics.
	}
	\label{fig:bifurcation_vary_k}
\end{figure}

\paragraph{The uncoupled case ($c=0$)} of \eqref{eq:controlled} is shown in \figref{fig:bifurcation_uncoupled}
Since the radius equations are cubic, one obtains three solutions until the damping (introduced by $k$) becomes stronger than energy input and no nontrivial solution is possible.
The control term introduces a coupling ``through the backdoor'' into the second equation of the system via the term $c_1 R_0 \cos(\Delta \Theta)$.
So, when varying $k$, the bifurcation diagram of $R_0$ shows a plain pitchfork bifurcation, the one for $R_1$ shows a distorted version of the pitchfork bifurcation due to this quasi-coupling.
\paragraph{The coupled case ($c\ne0$)} is  visualized for varying parameter $c$ in \figref{fig:bifurcation_vary_c} and for varying $k$ in \figref{fig:bifurcation_vary_k}.
Note that even in the case of vanishing control $k$ and non-zero coupling $c$, there is an asymmetry between $R_0$ and $R_1$ that is mediated by the third equation when $\Delta\omega\ne0$.
In \figref{fig:bifurcation_vary_c}, one can see a transition from the uncontrolled system to the controlled, pushing the right bifurcation point to higher values of $c$ with increasing coupling $k$.
At the same time, stationary solutions for the low end of $c$ cease to exist somewhere between $k=0.04$ and $k=0.06$.
The exact mechanism of this transition is out of the scope of this publication and will done in subsequent work together with a full bifurcation and stability analysis of this system.

An understanding comes from the following argument: In a three-dimensional system with cubic terms one can obtain, in principle, more than three solutions.
Since we have no additional objective in the GP run, it is clear that the particular choice of $k$ in such a simple control term as $k\dot{x}_i$ is just a representative of a larger class  of control laws.
To fix it to a specific value, or to enforce a symmetric situation, one has to add the corresponding terms in the cost function.
Since the control term is asymmetric, with increasing $k$, one changes the bifurcation scenario from perfect to imperfect.
This qualitative behavior is found in all graphs shown hereafter.

The behavior of $R_0$ and $R_1$ with increasing coupling and fixed control can be also be understood with a close look at \figref{fig:bifurcation_vary_c}.
At highest values of $c$ where stationary solutions exist, the values of $R_0$ and $R_1$ are approximately equal.
As the coupling strength diminuishes, the two radii take increasingly different values as for low values of $c$, the coupling is dominated by the asymmetric terms with $k$.
As the control strength $k$ is increased, stationary solutions at higher coupling strenghts exist and this behavior becomes more and more strongly pronounced where high $c$ values dominate the coupling which is thus symmetric.

However, beyond a certain control, $k$, no synchronization can be found at all -- the sudden stop of the curves is no artifact, but rather a true dynamical effect, this is seen in \figref{fig:bifurcation_vary_k}.
The overall behavior is clearly correct as the benchmark graph $c=0$ shows that the solution is lost at $k=\alpha$.
A similar observation holds for \figref{fig:bifurcation_vary_c}, where stationary solutions of \eqref{eq:controlled} cease to exist above some critical value of $c$.

These observations explain also, why the simple control term $k\dot{x}_i$ can push dynamics from synchronized to de-synchronized: it adds a damping of one oscillator $i$ that desynchronizes the two oscillators.
The value can be read off of the figures \figref{fig:bifurcation_vary_k} for a given coupling strength.

The GP-algorithm yields a value of $k = 1$, clearly this is larger than the critical value of approximately $0.05$ (For $c=0.022$, see \tabref{tab:two_vdp_desync_setup}), read off \figref{fig:bifurcation_vary_k}.
Now, one may ask why this value is chosen and not another one $k>0.05$.
The reason is simple, but it is hidden in the problem formulation: The objective function only requires that synchronization be destroyed.
This is possible for many functions and in particular for many simple functions with complexity $2$.
Among them, the control function $k\dot{x}$ is particularly appealing due to its simplicity.
The algorithm is now free to choose any $k>0.05$ and so it does.
The value $1$ is probably appearing because it is the first guess for a constant in the constant-optimization step of the algorithm and because it satisfies the objective.

The mathematical analysis also allows fine-tuning the control by varying the control parameter $k$ e.g. when the coupling strength changes and certain characteristics of the solution are to be kept.
Such adjustments could for example be read off \figref{fig:bifurcation_2D} where the height lines of certain values of the solutions parameters are tracked in the $c-k$ plane.

As a consequence, one can conclude that a careful formulation of the objective helps in obtaining a unique answer.
In the following we will not comment further on these details.
They must be considered in any application of the method, though.
We demonstrated that the symbolic regression performed by the GP algorithm produces results which are \emph{interpretable and tractable} with mathematical methods, here within the framework of dynamical systems.
This allows the subsequent step of exactly understanding the implications of a particular choice of control and choosing the one that is best suited to the needs of the particular problem at hand.
Moreover, once the dynamics with included control have been understood, the method allows tweaks to the control term (like adjusting the value of $k$) while understanding what will be happening.
Neither the analytical interpretation of the control nor the possibility of tweaking the control would not have been possible for a control achieved by a neural network due to its black box nature.

\section{Discussion and Conclusion}
\label{sec:conc}
In this work we demonstrate the use of explainable MLC methods -symbolic regression by GP- for rigorous analysis. We found several control laws results with similar score lead to different results with respect to stability analysis. In a general context, this result means that we can automatize analysis of the top models with rigorous mathematical methods, where stability is just one among others. For the big questions, though, like climate change, vehicle optimization (e.g. fuel reduction or predictive, automatized maintenance ), autonomous drive and alike this is of primordial importance.

We analyzed explainable MLC using a well-known control problem in dynamical systems: coupled self-sustained oscillators which exhibit synchronized behaviour or desynchronized one, depending on the system parameters.
In a previous study, we applied our control approach to dynamical systems composed of networks of coupled oscillators, starting from two coupled van der Pol oscillators up to a hierarchical network consisting of a few hundred oscillators. In this work, we reproduced and used these results for a subsequent stability analysis.
Such rigorous analyses are definitely not accessible using black-box or qualitative methods like, e.g., neural networks. 
The comparatively complex handling of GP in comparison with other symbolic methods like generalized regression is paid off if explainable solutions are needed. Due to the evolutionary nature of the method, it is not guaranteed that the global optimum is found, consequently a subsequent rigorous analysis is of great value.

As a result we find terms of different complexity leading to different levels of synchronization control, where synchronization is measured using the Kuramoto parameter. In both cases, synchronization and desynchronisaton, the found control laws are tested for stability. In Section~\ref{sec:bifurcation-analysis}, we demonstrate the potential of the methods by following the solutions numerically. We do not even touch a detailed analysis of eigenvalues and Lyapunov exponents which play a very important role in the dynamics of dynamical systems in general. 

Current efforts go to an automatization of this analysis following an explainable method, as well the fast sparse methods of generalized regression. To this end we plan to extend our existing framework Glyph. Using this kind of analysis will extend the range of explainable MLC to result in robust and interpretable control laws, a fact which is a unique selling point for explainable MLC. Clearly, this can end in a round trip where better objective functions are designed, taking into consideration the rigorous analysis and possibly prior domain knowledge, e.g. in the form of additive symmetry terms. 
In conclusion, we state that in terms of mathematical rigor, versatility and adaptability, the crystal-box method of GP is superior to other rather black-box methods, as artificial neural networks or support vector machines.

\section*{Acknowledgements}
We thank A. Pikovsky for synchronization wisdom, M. Rosenblum for providing input with respect to an application to human brain dynamics.

\section{Appendix \label{sec:appendix}}

In this section, we  give a brief summary of the implementation details and the parameters used in our setup.
Hyperparameters have been chosen empirically such that they lead to plausible and interpretable results on the chosen set of examples. We did not optimize the hyperparameters for convergence.

Our software is based on Glyph, a package developed by ourselves \cite{glyph}, which in turn uses other, standard python packages, e.g., constant optimization is conducted using the Levenberg-Marquardt least squares algorithm (\textit{scipy}) and numerical integration using the dopri5 solver (also \textit{scipy}). Random numbers are generated using the Mersenne Twister pseudo-random number generator provided by the \textit{random} module \cite{Matsumoto1998MersenneTwister}. Finally, the \textit{sympy} module is used for the simplification of symbolic mathematical expressions generated from the GP runs \cite{sympy2016}, for more details see \cite{glyph}.

\tabref{tab:setup} gives an overview of the methods and parameters used for the GP runs. Actual implementations can be found under the same name in the \textit{deap} module.

\begin{table}[h]
	\caption{General setup of the GP runs. \label{tab:setup}}
	\centering
	\begin{tabular}{ll}
		\toprule
		\addlinespace
		Function set & $\{+, -, \cdot, \sin, \cos, \exp\}$ \\  
		Population size & $500$ \\
		Max. generations & $20$\\
		MOO algorithm & NSGA-II \\
		\addlinespace
		Tree generation & \textit{halfandhalf} \\
		Min. height & 1 \\
		Max. height & 4 \\
		\addlinespace
		Selection & \textit{selTournament} \\
		Tournament size & $2$ \\
		\addlinespace
		Breeding & \textit{varOr} \\
		\addlinespace
		Recombination & \textit{cxOnePoint} \\
		Crossover probability & $0.5$ \\
		Crossover max. height & $20$ \\
		\addlinespace
		Mutation & \textit{mutUniform} \\
		Mutation probability & $0.2$ \\
		Mutation max. height & $20$ \\
		\addlinespace
		Constant optimization & \textit{leastsq} \\
		\addlinespace
		\bottomrule
	\end{tabular}
\end{table}

The following tables  \ref{tab:two_vdp_sync_setup} and \ref{tab:two_vdp_desync_setup} list  the setup used for two coupled oscillators forced to to synchronization and de-synchronization, respectively.
\begin{table}[htpb]
	\caption[Two coupled oscillators sync: setup]{Two Coupled Oscillators: System setup for forced synchronization.\label{tab:two_vdp_sync_setup}}
	\centering
	\begin{tabular}{lllll}
		\toprule
		\multicolumn{2}{c}{dynamic system} && \multicolumn{2}{c}{GP} \\
		\cmidrule{1-2} \cmidrule{4-5}
		\addlinespace
		$\omega_0$           & $\ln(4)$                         && cost functionals & $|\Omega_0 - \Omega_1|$ \\
		$\omega_1$           & $\ln(4) + 0.04$                  &&                  & length$(u)$ \\
		$\alpha,\,\beta,\,c$ & $0.1,\,1,\,0.022$                && argument set     & $\{\dot{x}_0, \dot{x}_1\}$ \\
		$\vec{x}(t_0)$       & $(1,1)$                          && constant set     & $\{k\}$ \\
		$\dot{\vec{x}}(t_0)$ & $(0,0)$                          && seed             & $3464542173339676227$ \\
		$t_0,\,t_n$          & $0,\,2000\tfrac{2\pi}{\omega_0}$ && \\
		$n$                  & $40000$                          && \\
		\addlinespace
		\bottomrule
	\end{tabular}
\end{table}

\begin{table}[htpb]
	\caption[Two coupled oscillators de-sync: setup]{Two Coupled Oscillators: System setup for forced de-synchronization.\label{tab:two_vdp_desync_setup}}
	\centering
	\begin{tabular}{lllll}
		\toprule
		\multicolumn{2}{c}{dynamic system} && \multicolumn{2}{c}{GP} \\
		\cmidrule{1-2} \cmidrule{4-5}
		\addlinespace
		$\omega_0$           & $\ln(4)$                         && cost functionals & $\exp(-|\Omega_0 - \Omega_1|)$ \\
		$\omega_1$           & $\ln(4) + 0.015$                 &&                  & length$(u)$ \\
		$\alpha,\,\beta,\,c$ & $0.1,\,1,\,0.022$                && argument set     & $\{\dot{x}_0, \dot{x}_1\}$ \\
		$\vec{x}(t_0)$       & $(1,1)$                          && constant set     & $\{k\}$ \\
		$\dot{\vec{x}}(t_0)$ & $(0,0)$                          && seed             & $2590675513212712687$ \\
		$t_0,\,t_n$          & $0,\,2000\tfrac{2\pi}{\omega_0}$ && \\
		$n$                  & $40000$                          && \\
		\addlinespace
		\bottomrule
	\end{tabular}
\end{table}


\begin{thebibliography}{10}
	\expandafter\ifx\csname url\endcsname\relax
	\def\url#1{\texttt{#1}}\fi
	\expandafter\ifx\csname urlprefix\endcsname\relax\def\urlprefix{URL }\fi
	\expandafter\ifx\csname href\endcsname\relax
	\def\href#1#2{#2} \def\path#1{#1}\fi
	
	\bibitem{10.1007/978-3-319-99740-7_21}
	R.~Goebel, A.~Chander, K.~Holzinger, F.~Lecue, Z.~Akata, S.~Stumpf,
	P.~Kieseberg, A.~Holzinger, Explainable ai: The new 42?, in: A.~Holzinger,
	P.~Kieseberg, A.~M. Tjoa, E.~Weippl (Eds.), Machine Learning and Knowledge
	Extraction, Springer International Publishing, Cham, 2018, pp. 295--303
	(2018).
	
	\bibitem{PhysRevLett.64.1196}
	E.~Ott, C.~Grebogi, J.~A. Yorke,
	\href{https://doi.org/10.1103/physrevlett.64.1196}{Controlling chaos}, Phys.
	Rev. Lett. 64~(11) (1990) 1196--1199 (Mar. 1990).
	\newblock \href {https://doi.org/10.1103/physrevlett.64.1196}
	{\path{doi:10.1103/physrevlett.64.1196}}.
	\newline\urlprefix\url{https://doi.org/10.1103/physrevlett.64.1196}
	
	\bibitem{chen2003chaos}
	\href{https://doi.org/10.1007/b79666}{Chaos Control}, Springer Berlin
	Heidelberg, 2003 (2003).
	\newblock \href {https://doi.org/10.1007/b79666} {\path{doi:10.1007/b79666}}.
	\newline\urlprefix\url{https://doi.org/10.1007/b79666}
	
	\bibitem{haken2006brain}
	H.~Haken, \href{https://books.google.de/books?id=8elDAAAAQBAJ}{Brain Dynamics:
		{Synchronization} and Activity Patterns in Pulse-Coupled Neural Nets with
		Delays and Noise}, Springer Series in Synergetics, Springer Berlin
	Heidelberg, 2006 (2006).
	\newline\urlprefix\url{https://books.google.de/books?id=8elDAAAAQBAJ}
	
	\bibitem{schwalb2008history}
	J.~M. Schwalb, C.~Hamani, \href{https://doi.org/10.1016/j.nurt.2007.11.003}{The
		history and future of deep brain stimulation}, Neurotherapeutics 5~(1) (2008)
	3--13 (Jan. 2008).
	\newblock \href {https://doi.org/10.1016/j.nurt.2007.11.003}
	{\path{doi:10.1016/j.nurt.2007.11.003}}.
	\newline\urlprefix\url{https://doi.org/10.1016/j.nurt.2007.11.003}
	
	\bibitem{Pikovsky2003Synchronization}
	A.~Pikovsky, M.~Rosenblum, J.~Kurths,
	\href{https://doi.org/10.1017/cbo9780511755743}{Synchronization}, Cambridge
	University Press, 2001 (2001).
	\newblock \href {https://doi.org/10.1017/cbo9780511755743}
	{\path{doi:10.1017/cbo9780511755743}}.
	\newline\urlprefix\url{https://doi.org/10.1017/cbo9780511755743}
	
	\bibitem{strogatz2003sync}
	S.~H. Strogatz, Sync: {How} order emerges from chaos in the universe, nature,
	and daily life, Hyperion, 2003 (2003).
	
	\bibitem{Gout2018}
	J.~Gout, M.~Quade, K.~Shafi, R.~K. Niven, M.~Abel,
	\href{https://doi.org/10.1007/s11071-017-3925-z}{Synchronization control of
		oscillator networks using symbolic regression}, Nonlinear Dynamics 91~(2)
	(2018) 1001--1021 (Jan 2018).
	\newblock \href {https://doi.org/10.1007/s11071-017-3925-z}
	{\path{doi:10.1007/s11071-017-3925-z}}.
	\newline\urlprefix\url{https://doi.org/10.1007/s11071-017-3925-z}
	
	\bibitem{yang1997impulsive}
	\href{https://doi.org/10.1109/81.633887}{Impulsive stabilization for control
		and synchronization of chaotic systems: {Theory} and application to secure
		communication}, IEEE Trans. Circuits Syst. I 44~(10) (1997) 976--988 (1997).
	\newblock \href {https://doi.org/10.1109/81.633887}
	{\path{doi:10.1109/81.633887}}.
	\newline\urlprefix\url{https://doi.org/10.1109/81.633887}
	
	\bibitem{li2011adaptive}
	Z.~Li, X.~Cao, N.~Ding,
	\href{https://doi.org/10.1109/tfuzz.2011.2143417}{Adaptive fuzzy control for
		synchronization of nonlinear teleoperators with stochastic time-varying
		communication delays}, IEEE Trans. Fuzzy Syst. 19~(4) (2011) 745--757 (Aug.
	2011).
	\newblock \href {https://doi.org/10.1109/tfuzz.2011.2143417}
	{\path{doi:10.1109/tfuzz.2011.2143417}}.
	\newline\urlprefix\url{https://doi.org/10.1109/tfuzz.2011.2143417}
	
	\bibitem{shokri2014optimal}
	H.~Shokri-Ghaleh, A.~Alfi,
	\href{https://doi.org/10.1007/s11071-014-1589-5}{Optimal synchronization of
		teleoperation systems via cuckoo optimization algorithm}, Nonlinear Dyn
	78~(4) (2014) 2359--2376 (Aug. 2014).
	\newblock \href {https://doi.org/10.1007/s11071-014-1589-5}
	{\path{doi:10.1007/s11071-014-1589-5}}.
	\newline\urlprefix\url{https://doi.org/10.1007/s11071-014-1589-5}
	
	\bibitem{hammond2007pathological}
	C.~Hammond, H.~Bergman, P.~Brown,
	\href{https://doi.org/10.1016/j.tins.2007.05.004}{Pathological
		synchronization in {Parkinson's} disease: {Networks,} models and treatments},
	Trends in Neurosciences 30~(7) (2007) 357--364 (Jul. 2007).
	\newblock \href {https://doi.org/10.1016/j.tins.2007.05.004}
	{\path{doi:10.1016/j.tins.2007.05.004}}.
	\newline\urlprefix\url{https://doi.org/10.1016/j.tins.2007.05.004}
	
	\bibitem{dorfler2013synchronization}
	F.~Dorfler, M.~Chertkov, F.~Bullo,
	\href{https://doi.org/10.1073/pnas.1212134110}{Synchronization in complex
		oscillator networks and smart grids}, Proceedings of the National Academy of
	Sciences 110~(6) (2013) 2005--2010 (Jan. 2013).
	\newblock \href {https://doi.org/10.1073/pnas.1212134110}
	{\path{doi:10.1073/pnas.1212134110}}.
	\newline\urlprefix\url{https://doi.org/10.1073/pnas.1212134110}
	
	\bibitem{kirk2012optimal}
	D.~E. Kirk, Optimal control theory: {An} introduction, Courier Corporation,
	2012 (2012).
	
	\bibitem{nocedal2006numerical}
	\href{https://doi.org/10.1007/b98874}{Numerical Optimization}, Springer-Verlag,
	1999 (1999).
	\newblock \href {https://doi.org/10.1007/b98874} {\path{doi:10.1007/b98874}}.
	\newline\urlprefix\url{https://doi.org/10.1007/b98874}
	
	\bibitem{opt4ml}
	S.~Sra, S.~Nowozin, S.~J. Wright, Optimization for Machine Learning, MIT Press,
	Cambridge, USA, 2011 (2011).
	
	\bibitem{Becerra2008OptimalControl}
	V.~Becerra, \href{https://doi.org/10.4249/scholarpedia.5354}{Optimal control},
	Scholarpedia 3~(1) (2008) 5354 (2008).
	\newblock \href {https://doi.org/10.4249/scholarpedia.5354}
	{\path{doi:10.4249/scholarpedia.5354}}.
	\newline\urlprefix\url{https://doi.org/10.4249/scholarpedia.5354}
	
	\bibitem{Kirk1970OptimalControl}
	\href{https://doi.org/10.1007/0-387-29903-3}{Optimal Control Theory},
	Springer-Verlag, 2000 (2000).
	\newblock \href {https://doi.org/10.1007/0-387-29903-3}
	{\path{doi:10.1007/0-387-29903-3}}.
	\newline\urlprefix\url{https://doi.org/10.1007/0-387-29903-3}
	
	\bibitem{scholl2008handbook}
	\href{https://doi.org/10.1002/9783527622313}{Handbook of Chaos Control},
	Wiley-VCH Verlag GmbH \& Co. KGaA, 2007 (Oct. 2007).
	\newblock \href {https://doi.org/10.1002/9783527622313}
	{\path{doi:10.1002/9783527622313}}.
	\newline\urlprefix\url{https://doi.org/10.1002/9783527622313}
	
	\bibitem{Koza1992GeneticProgramming}
	J.~R. Koza, Genetic Programming: {On} the Programming of Computers by Means of
	Natural Selection, MIT Press, 1992 (1992).
	
	\bibitem{schmidt2009distilling}
	M.~Schmidt, H.~Lipson,
	\href{https://doi.org/10.1126/science.1165893}{Distilling free-form natural
		laws from experimental data}, Science 324~(5923) (2009) 81--85 (Apr. 2009).
	\newblock \href {https://doi.org/10.1126/science.1165893}
	{\path{doi:10.1126/science.1165893}}.
	\newline\urlprefix\url{https://doi.org/10.1126/science.1165893}
	
	\bibitem{vladislavleva2009order}
	E.~Vladislavleva, G.~Smits, D.~den Hertog,
	\href{https://doi.org/10.1109/tevc.2008.926486}{Order of nonlinearity as a
		complexity measure for models generated by symbolic regression via {Pareto}
		genetic programming}, IEEE Trans. Evol. Computat. 13~(2) (2009) 333--349
	(Apr. 2009).
	\newblock \href {https://doi.org/10.1109/tevc.2008.926486}
	{\path{doi:10.1109/tevc.2008.926486}}.
	\newline\urlprefix\url{https://doi.org/10.1109/tevc.2008.926486}
	
	\bibitem{Quade2016Prediction}
	M.~Quade, M.~Abel, K.~Shafi, R.~K. Niven, B.~R. Noack,
	\href{https://doi.org/10.1103/physreve.94.012214}{Prediction of dynamical
		systems by symbolic regression}, Phys. Rev. E 94~(1) (Jul. 2016).
	\newblock \href {https://doi.org/10.1103/physreve.94.012214}
	{\path{doi:10.1103/physreve.94.012214}}.
	\newline\urlprefix\url{https://doi.org/10.1103/physreve.94.012214}
	
	\bibitem{shokri2014comparison}
	H.~Shokri-Ghaleh, A.~Alfi, \href{https://doi.org/10.1016/j.asoc.2014.07.020}{A
		comparison between optimization algorithms applied to synchronization of
		bilateral teleoperation systems against time delay and modeling
		uncertainties}, Applied Soft Computing 24 (2014) 447--456 (Nov. 2014).
	\newblock \href {https://doi.org/10.1016/j.asoc.2014.07.020}
	{\path{doi:10.1016/j.asoc.2014.07.020}}.
	\newline\urlprefix\url{https://doi.org/10.1016/j.asoc.2014.07.020}
	
	\bibitem{markus_quade_2017_801819}
	M.~Quade, J.~Gout, M.~Abel,
	\href{https://doi.org/10.5281/zenodo.801819}{Ambrosys/glyph: {V0.3.2}} (Jun.
	2017).
	\newblock \href {https://doi.org/10.5281/zenodo.801819}
	{\path{doi:10.5281/zenodo.801819}}.
	\newline\urlprefix\url{https://doi.org/10.5281/zenodo.801819}
	
	\bibitem{doi:10.2514/6.2018-3684}
	Y.~E.~S. M., P.~Oswald, S.~Sattler, P.~Kumar, R.~Radespiel, C.~Behr,
	M.~Sinapius, J.~Petersen, P.~Wierach, M.~Quade, M.~Abel, B.~R. Noack,
	R.~Semaan, \href{https://arc.aiaa.org/doi/abs/10.2514/6.2018-3684}{Open- and
		closed-loop control investigations of unsteady Coanda actuation on a
		high-lift configuration}.
	\newblock \href
	{http://arxiv.org/abs/https://arc.aiaa.org/doi/pdf/10.2514/6.2018-3684}
	{\path{arXiv:https://arc.aiaa.org/doi/pdf/10.2514/6.2018-3684}}, \href
	{https://doi.org/10.2514/6.2018-3684} {\path{doi:10.2514/6.2018-3684}}.
	\newline\urlprefix\url{https://arc.aiaa.org/doi/abs/10.2514/6.2018-3684}
	
	\bibitem{Strogatz2006SyncBasin}
	D.~A. Wiley, S.~H. Strogatz, M.~Girvan,
	\href{https://doi.org/10.1063/1.2165594}{The size of the sync basin}, Chaos
	16~(1) (2006) 015103 (Mar. 2006).
	\newblock \href {https://doi.org/10.1063/1.2165594}
	{\path{doi:10.1063/1.2165594}}.
	\newline\urlprefix\url{https://doi.org/10.1063/1.2165594}
	
	\bibitem{AUTO07p}
	E.~J. Doedel, T.~F. Fairgrieve, B.~Sandstede, A.~R. Champneys, Y.~A. Kuznetsov,
	X.~Wang, Auto-07p: Continuation and bifurcation software for ordinary
	differential equations, Tech. rep. (2007).
	
	\bibitem{goldstine2012history}
	H.~H. Goldstine, \href{https://doi.org/10.1007/978-1-4613-8106-8}{A History of
		the Calculus of Variations from the 17th through the 19th Century}, Springer
	New York, 1980 (1980).
	\newblock \href {https://doi.org/10.1007/978-1-4613-8106-8}
	{\path{doi:10.1007/978-1-4613-8106-8}}.
	\newline\urlprefix\url{https://doi.org/10.1007/978-1-4613-8106-8}
	
	\bibitem{press2007numerical}
	W.~H. Press, Numerical recipes 3rd edition: {The} art of scientific computing,
	Cambridge university press, 2007 (2007).
	
	\bibitem{Lewis12995OptimalControl}
	F.~L. Lewis, D.~L. Vrabie, V.~L. Syrmos,
	\href{https://doi.org/10.1002/9781118122631}{Optimal Control}, John Wiley \&
	Sons, Inc., 2012 (Jan. 2012).
	\newblock \href {https://doi.org/10.1002/9781118122631}
	{\path{doi:10.1002/9781118122631}}.
	\newline\urlprefix\url{https://doi.org/10.1002/9781118122631}
	
	\bibitem{Bryson1975OptimalControl}
	A.~E. {Bryson (Jr)}, Y.~Ho, Applied Optimal Control, Halsted Press, 1975
	(1975).
	
	\bibitem{Athans2006OptimalControl}
	M.~Athans, P.~L. Falb, Optimal Control: {An} Introduction to the Theory and Its
	Applications, Dover Publications, 2006 (2006).
	
	\bibitem{AHNERT2007764}
	K.~Ahnert, M.~Abel,
	\href{http://www.sciencedirect.com/science/article/pii/S0010465507003116}{Numerical
		differentiation of experimental data: local versus global methods}, Computer
	Physics Communications 177~(10) (2007) 764 -- 774 (2007).
	\newblock \href {https://doi.org/https://doi.org/10.1016/j.cpc.2007.03.009}
	{\path{doi:https://doi.org/10.1016/j.cpc.2007.03.009}}.
	\newline\urlprefix\url{http://www.sciencedirect.com/science/article/pii/S0010465507003116}
	
	\bibitem{Gene2014FeedbackControl}
	G.~F. Franklin, J.~D. Powell, A.~Emami-Naeini, Feedback Control of Dynamic
	Systems, 7th Edition, Pearson, 2014 (2014).
	
	\bibitem{PhysRevLett.83.3422}
	H.~U. Voss, P.~Kolodner, M.~Abel, J.~Kurths,
	\href{https://link.aps.org/doi/10.1103/PhysRevLett.83.3422}{Amplitude
		equations from spatiotemporal binary-fluid convection data}, Phys. Rev. Lett.
	83 (1999) 3422--3425 (Oct 1999).
	\newblock \href {https://doi.org/10.1103/PhysRevLett.83.3422}
	{\path{doi:10.1103/PhysRevLett.83.3422}}.
	\newline\urlprefix\url{https://link.aps.org/doi/10.1103/PhysRevLett.83.3422}
	
	\bibitem{PhysRevE_57_2820}
	H.~Voss, M.~J. B\"unner, M.~Abel,
	\href{https://link.aps.org/doi/10.1103/PhysRevE.57.2820}{Identification of
		continuous, spatiotemporal systems}, Phys. Rev. E 57 (1998) 2820--2823 (Mar
	1998).
	\newblock \href {https://doi.org/10.1103/PhysRevE.57.2820}
	{\path{doi:10.1103/PhysRevE.57.2820}}.
	\newline\urlprefix\url{https://link.aps.org/doi/10.1103/PhysRevE.57.2820}
	
	\bibitem{Cramer1985Representation}
	N.~L. Cramer, A representation for the adaptive generation of simple sequential
	programs, in: J.~J. Grefenstette (Ed.), Proceedings of an International
	Conference on Genetic Algorithms and the Applications, Psychology Press,
	1985, pp. 183--187 (1985).
	
	\bibitem{Poli2008FieldGuide}
	\href{https://doi.org/10.1007/3-540-48885-5}{Genetic Programming}, Springer
	Berlin Heidelberg, 1999 (1999).
	\newblock \href {https://doi.org/10.1007/3-540-48885-5}
	{\path{doi:10.1007/3-540-48885-5}}.
	\newline\urlprefix\url{https://doi.org/10.1007/3-540-48885-5}
	
	\bibitem{Yang2011Metaheuristics}
	X.-S. Yang, \href{https://doi.org/10.4249/scholarpedia.11472}{Metaheuristic
		optimization}, Scholarpedia 6~(8) (2011) 11472 (2011).
	\newblock \href {https://doi.org/10.4249/scholarpedia.11472}
	{\path{doi:10.4249/scholarpedia.11472}}.
	\newline\urlprefix\url{https://doi.org/10.4249/scholarpedia.11472}
	
	\bibitem{glyph}
	M.~A. M~Quade, J~Gout, Glyph: Symbolic regression tools, Journal of Open
	Research Software 1~(7).
	
	\bibitem{Kuramoto1975}
	Y.~Kuramoto, Lecture notes in physics, in: H.~Araki (Ed.), International
	Symposium on Mathematical Problems in Theoretical Physics, Vol.~39, Springer,
	1975, p. 420 (1975).
	
	\bibitem{Kuramoto1984}
	Y.~Kuramoto, \href{https://doi.org/10.1007/978-3-642-69689-3}{Chemical
		Oscillations, Waves, and Turbulence}, Springer Berlin Heidelberg, 1984
	(1984).
	\newblock \href {https://doi.org/10.1007/978-3-642-69689-3}
	{\path{doi:10.1007/978-3-642-69689-3}}.
	\newline\urlprefix\url{https://doi.org/10.1007/978-3-642-69689-3}
	
	\bibitem{Cohen1994TimeFrequencyAnalysis}
	L.~Cohen, Time Frequency Analysis: {Theory} and Applications, 1st Edition,
	Prentice Hall, 1994 (1994).
	
	\bibitem{Strogatz2014NLD}
	S.~H. Strogatz, Nonlinear Dynamics And Chaos, 2nd Edition, Westview Press, 2015
	(2015).
	
	\bibitem{Matsumoto1998MersenneTwister}
	M.~Matsumoto, T.~Nishimura,
	\href{https://doi.org/10.1145/272991.272995}{Mersenne twister: {A}
		623-dimensionally equidistributed uniform pseudo-random number generator},
	ACM Trans. Model. Comput. Simul. 8~(1) (1998) 3--30 (Jan. 1998).
	\newblock \href {https://doi.org/10.1145/272991.272995}
	{\path{doi:10.1145/272991.272995}}.
	\newline\urlprefix\url{https://doi.org/10.1145/272991.272995}
	
	\bibitem{sympy2016}
	{SymPy Dev Team}, \href{http://www.sympy.org}{{{SymPy}:} {Python} library for
		symbolic mathematics} (2016).
	\newline\urlprefix\url{http://www.sympy.org}
	
\end{thebibliography}
\end{document}